\documentclass{article}

     \usepackage[final]{neurips_2022}

\usepackage[utf8]{inputenc} 
\usepackage[T1]{fontenc}    
\usepackage{hyperref}       
\usepackage{url}            
\usepackage{booktabs}       
\usepackage{amsfonts}       
\usepackage{nicefrac}       
\usepackage{microtype}      
\usepackage{xcolor}         
\usepackage{smile}
\DeclareMathOperator*{\argtopm}{arg\,top-M}
\usepackage{enumitem}
\usepackage{wrapfig}

\newcommand{\eg}{\textit{e.g.}}
\usepackage{listings}

\title{Visual Clues: Bridging Vision and Language Foundations for Image Paragraph Captioning}

\author{%
Yujia Xie, Luowei Zhou, Xiyang Dai, Lu Yuan, Nguyen Bach, Ce Liu, Michael Zeng \\ Microsoft
\\ \texttt{\small \{yujiaxie, luowei.zhou, xiyang.dai, luyuan, nguyenbach, ce.liu, nzeng\}\@microsoft.com} 
}

\begin{document}

\maketitle

\begin{abstract}



People say, ``\textit{A picture is worth a thousand words}''. Then how can we get the rich information out of the image? We argue that by using \textit{visual clues} to bridge large pretrained vision foundation models and language models, we can do so without any extra cross-modal training. Thanks to the strong zero-shot capability of foundation models, we start by constructing a rich semantic representation of the image (\eg, image tags, object attributes / locations, captions) as a structured textual prompt, called \textit{visual clues}, using a vision foundation model. Based on visual clues, we use large language model to produce a series of comprehensive descriptions for the visual content, which is then verified by the vision model again to select the candidate that aligns best with the image. We evaluate the quality of generated descriptions by quantitative and qualitative measurement. The results demonstrate the effectiveness of such a structured semantic representation.

\end{abstract}
\section{Introduction}

\begin{figure}[h]
  \centering
  \fbox{
    \begin{minipage}[t]{0.3\linewidth}
    \includegraphics[height=5\baselineskip]{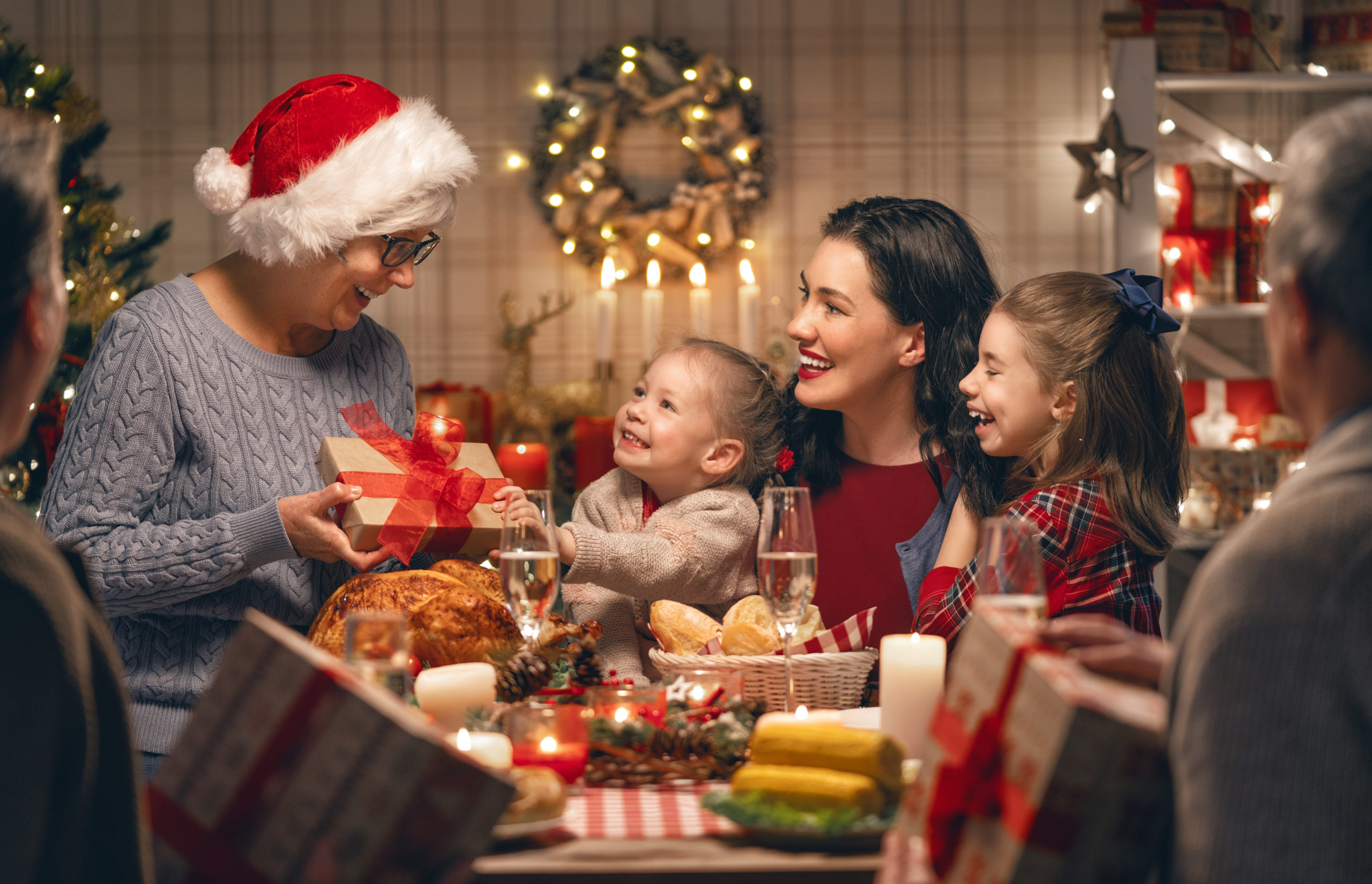}
    \end{minipage}
    \hspace{-10pt}
    \begin{minipage}[t]{0.6\linewidth}
    \vspace{-53pt}
    \textit{This image is of a family celebrating Christmas. They are all gathered around a dinner table, with a turkey and other food on it. The family is smiling and seems to be enjoying themselves. There is a Christmas tree in the background and some Christmas lights on the walls.}
    \end{minipage}
    }
    \fbox{
    \begin{minipage}[t]{0.3\linewidth}
    \includegraphics[height=5\baselineskip]{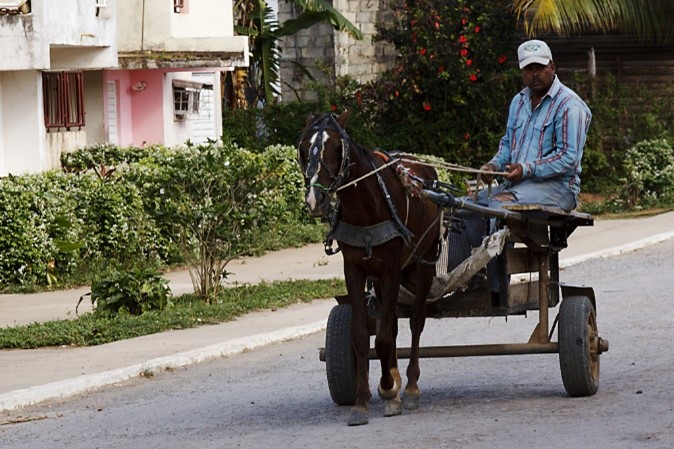}
    \end{minipage}
    \hspace{-10pt}
    \begin{minipage}[t]{0.6\linewidth}
    \vspace{-53pt}
    \textit{This image features a horse and buggy travelling down a road in the town of Holguin, Cuba. The horse is harnessed to the buggy and is pulling it along, while a farmworker rides behind. This image captures the everyday life of Cubans, with their traditional horse-drawn carts still in use.}
    \end{minipage}
    }
  \caption{\label{fig:illu} Examples of generated image paragraph.}
\end{figure}

\begin{quote}
    ``\textit{Vision is a process that produces from images of the external world a \textbf{description} that is useful to the viewer and not cluttered with irrelevant information.}''\\
    \mbox{}\hfill---David Marr, \textit{Vision, p31}
\end{quote}

What makes a good ``\textit{description}'' for vision? Over the past several decades, computer vision pioneers drew inspiration from neural science, cognitive science, and psychophysics~\citep{marr2010vision}, pointing us to the North Stars~\citep{fei2022searching}, some among them being image classification and object detection. Despite the tremendous progress that has been made, much of these object-centric works remain a proxy for an eventual task or application that requires a holistic view of the visual content, involving concepts beyond objects: actions, attributes, and relations, to name a few. 

In our work, we argue that textual representation suffices such ``description''. It brings forth a more holistic visual representation than categorical labels. It allows machines to interpret visual signals through descriptive captions~\citep{zhou2020unified, li2022blip}, and perform more language-heavy tasks such as question-answers~\citep{rajpurkar2016squad}, or multi-round dialogues~\citep{li2017dailydialog}. On the other hand, the access to abundant web multimodal language data (\eg, image alt-text, video subtitles) provides us with the fuel for powering neural visual representations from contrastive language-image pre-training (CLIP, \citet{yuan2021florence,radford2021learning}). The marriage of the two renders a new computer vision system that is faithful, generic, and versatile.

We name this new computer vision system \textbf{BEST}, for Bridging with Explicit Structured Textural clues. We start by constructing a semantic representation of the image. This semantic representation, which we referred to as \textit{visual clues}, comprises rich semantic components, from object and attribute tags to localized detection regions and region captions. Powered by the recent advances in vision foundation model Florence \citep{yuan2021florence}, the visual clues are rich in open-vocabulary expressions, marking a major difference compared to existing symbolic approaches (\eg, scene graphs~\cite{krishna2017visual}) with closed-set vocabularies.

The visual clues are interpretable, not only for humans, but also for machines. Take the generative language model GPT-3~\citep{brown2020language}. The visual clues could be digested by GPT-3, which in return produces crisp language descriptions that are sensible to the viewer while not cluttered with irrelevant information from the visual clues. Whereas this open-loop process could potentially suffer from object hallucination issues \citep{maynez2020faithfulness, zhou2020detecting} as the outputs from GPT-3 are not governed by any means, we further deploy a closed-loop verification procedure that grounds descriptions back to the original image.

To evaluate the quality of the language descriptions, we resort to an existing task named Image Paragraph Captioning (IPC), but with a twist. IPC aims to address the demand for generating long, coherent, and informative descriptions of the whole visual content of an image \citep{krause2017hierarchical},
which can eventually be used for many applications including poetry composition \citep{liu2018beyond}, automatic recipe generation \citep{salvador2019inverse}, visual storytelling \citep{huang2016visual}, advertisement generation, or help blind or visually-impaired people see better.
The existing metrics for IPC such as BLEU \citep{papineni2002bleu}, METEOR \citep{denkowski2014meteor}, and CIDEr \citep{vedantam2015cider} encourage exact matching between semantics in generated captions and those in the reference. However, they over penalize visual details that are not annotated thus compromising their qualifications for measuring overall representation quality.
Inspired by \citet{anderson2016spice, krishna2017visual}, we propose to measure the accuracy on \textit{scene graphs} extracted from generated text against human-annotated graphs, which, as suggested by \citet{anderson2016spice}, co-relates better with human judgment. 

The contributions are twofold. First, we propose a general framework for semantic visual representation and showcase its application to image paragraph captioning. The framework is simple yet highly extendable, allowing new components to be plug-in and supporting other use scenarios that require a holistic view of the visual content. Second, we benchmark the effectiveness of the proposed model on its capacity for representing visual concepts (\eg, scene graphs) and set new state-of-the-art results. 

\noindent\textbf{Notations.} We denote $\langle \cdot, \cdot \rangle$ as inner product between two vectors, $|\mathcal{A}|$ as the cardinality of set $\mathcal{A}$.

\section{Related Works}

\noindent\textbf{Image paragraph generation. }
The task of generating image paragraphs is first introduced by \citet{krause2017hierarchical}. Conditioned on the visual features, they first train a sentence recurrent neural network (RNN) to output sentence topics, and then feed each of the topics into another RNN to generate the paragraphs. 
\citet{liang2017recurrent} further improve the hierarchical RNN framework by introducing an adversarial discriminator for smoother transitions between sentences.
\citet{chatterjee2018diverse} also address cross-sentence topic consistency by a global coherence vector. 
\citet{melas2018training} add a repeat penalty to the optimization, to prevent the appearance of repeated sequences. 
\citet{wang2019convolutional} use convolutional auto-encoder for topic modeling based on region-level image features. 
Along this line, many other works have been done \citep{dai2017towards, luo2019curiosity, mao2018show, xu2020interactive, guo2021matching, shi2021s2td}. Most of the proposed models, however, are trained on \textit{Stanford image-paragraph dataset} \citep{krause2017hierarchical}, which only contains 14 thousand of training paragraphs for its expensive nature to collect. 
Due to lack of data, the generated paragraphs usually lack coherence both locally and globally. Therefore, many of the above works aim to make the best use of data to improve the coherence. 
Yet nowadays, large language models can generate long coherent paragraphs by default. Our work, leveraging recent progress of large pretrained models, focuses more on how to guide and constrain the generated text instead.















\noindent\textbf{Constrained text generation} In recent years, rapid progress has been made in vision-language pretraining (VLP). CLIP \citep{radford2021learning}, ALIGN \citep{jia2021scaling}, and Florence \citep{yuan2021florence} are proposed to encode vision and language into a \textit{joint} representation space for crossmodal alignment tasks, \eg, zero-shot image classification. Another line of research, \eg, SimVLM \citep{wang2021simvlm}, FLAVA \citep{singh2021flava}, BLIP \citep{li2022blip}, CoCa \citep{Yu2022CoCaCC} and many others \citep{cho2021unifying, wang2022unifying, zhu2021uni, alayrac2022flamingo} adopt encoder-decoder models trained with generative losses. Those models are capable of performing image captioning in a zero-shot manner. 
A concurrent work, Socratic Models (SM, \citet{zeng2022socratic}), also use textual data to bridge the domain gap between vision-language models and language models. The model, however, is stronger in retrieval tasks than captioning tasks as we will show later. 
There are also other works leveraging large language models to solve vision tasks, \eg, PICa \citep{yang2021empirical} uses GPT-3 \citep{brown2020language} to extract commonsense knowledge for visual question answering tasks, MAGIC \citep{su2022language} uses a CLIP-induced score to regularize the language generation  so that it is semantically related to the given image, and VisualGPT \citep{chen2022visualgpt} employs a self-resurrecting encoder-decoder attention mechanism to adapt the language models with a small amount of in-domain image-text data.



\section{Framework}

\begin{figure}
  \centering  \includegraphics[scale=0.5]{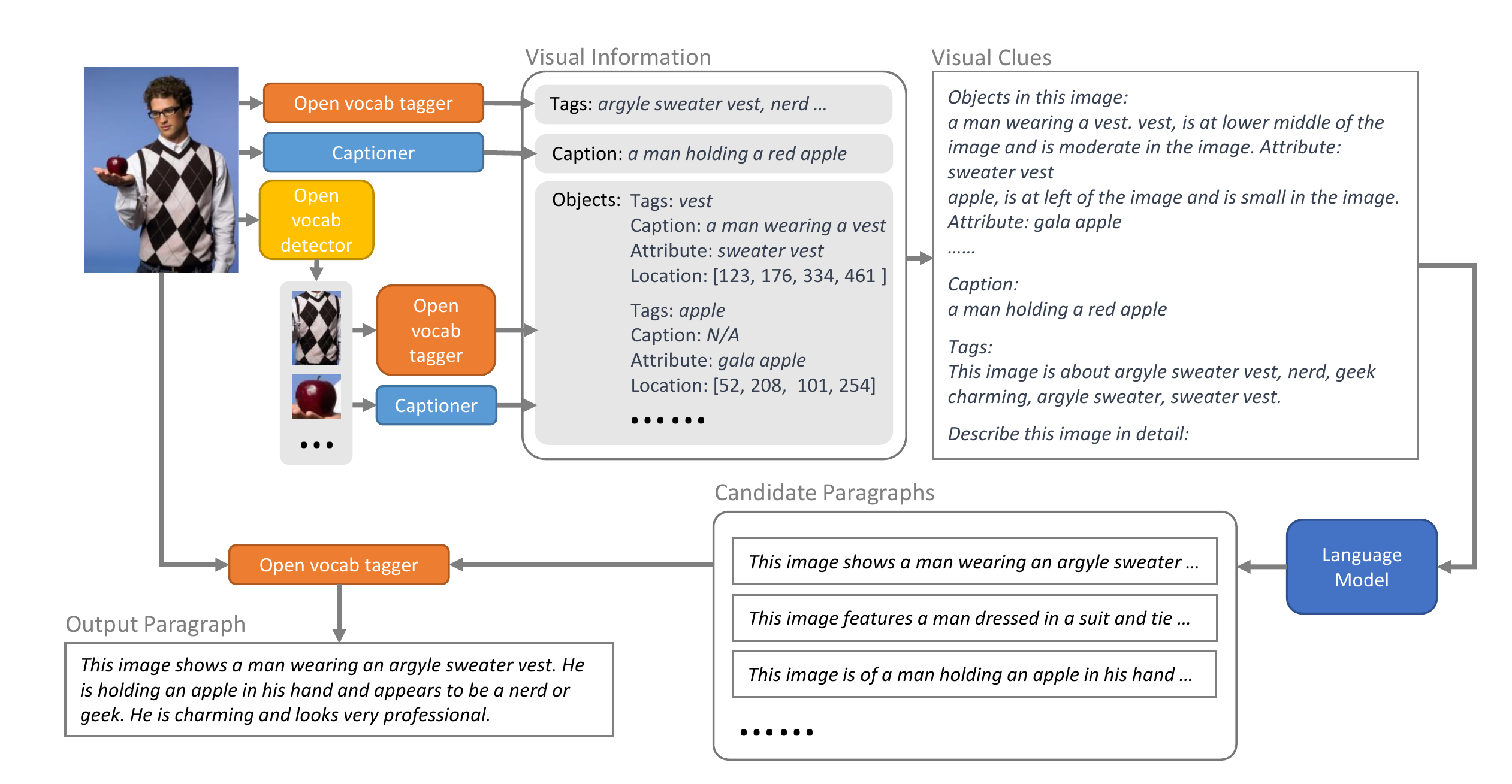}
  \caption{\label{fig:framwork} Framework demonstration. The orange \textit{open vocab tagger} box corresponds to the image encoder $f_\text{v}(\cdot)$ and the text encoder as $f_\text{t}(\cdot)$. The blue \textit{captioner} box is the caption model $c(\cdot)$.}
\end{figure}

Given an image $I$, our goal is to generate long and coherent descriptive text based on image inputs, leveraging only the existing pretrained models. Our framework can be divided into three stages:
\begin{enumerate}
\itemsep0em 
    \item Represent $I$ with visual clues $S$, which contain the rich visual information;
    \item Feed the visual clues into a language model to generate $K$ candidate paragraphs $\{T_i\}_{i=1}^K$;
    \item Select the best paragraph $T^*$ from the candidates $\{T_i\}_{i=1}^K$.
\end{enumerate}
The overall framework is illustrated in Figure \ref{fig:framwork}. We will then elaborate on each of them. 

\subsection{Visual Clue Extraction}

We leverage three state-of-the-art models with the open-vocabulary capability to extract the visual information, namely, the concise tags, the short captions, and the local descriptions. 

\textbf{Concise tags.} The first model we use is the contrastively trained vision-language models, \eg, CLIP \citep{radford2021learning}, Florence \citep{yuan2021florence}. Such models are pretrained on image-text pairs $\{x_i, y_i\}$, and is composed of the image encoder $f_\text{v}(\cdot)$ and the text encoder $f_\text{t}(\cdot)$. Given a minibatch $\mathcal{B}$, the models are optimized by contrastive loss
\begin{equation*}
\mathcal{L} = - \frac{1}{|\mathcal{B}|} \sum_{x_i, y_i \in \mathcal{B}} \left(\frac{\exp (\langle f_\text{v}(x_i), f_\text{t}(y_i) \rangle / \tau)}{\sum_{y_j \in \mathcal{B}, j \neq i} \exp (\langle f_\text{v}(x_i), f_\text{t}(y_j) \rangle / \tau)} 
     + \frac{\exp (\langle f_\text{v}(x_i), f_\text{t}(y_i) \rangle / \tau)}{\sum_{x_j \in \mathcal{B}, j \neq i} \exp (\langle f_\text{v}(x_j), f_\text{t}(y_i) \rangle / \tau)}\right),
\end{equation*}
where $\tau$ is the temperature. This loss explicitly uses inner product $\langle \cdot, \cdot \rangle$ to measure the similarity between the encoded image $f_\text{v}(x_i)$ and encoded text $f_\text{t}(y_j)$, and higher similarities are encouraged if the images and texts are paired. Therefore, such a pretrained model is capable of selecting the tags that describe the image $I$ from a set of customized tags by computing the similarities. Given a set of tags $\{t_i\}_{i=1}^N$, we compute the similarities between the input image $I$ and the tags, and adopt the tags with top-$M$ similarities,
\begin{equation}
    \mathcal{T} = \{t^*_j\}_{j=1}^M = \argtopm_{t_i, i=1, \cdots, N}  \langle f_\text{v}(I), f_\text{t}(t_i) \rangle.
\end{equation}
\textbf{Short captions.} The second model is a caption model $c(\cdot)$. We use it to generate an overall image description $c(I)$. 

\textbf{Local descriptions.} The third model is an object detection model. We adopt a well-trained object detector, to provide us with the locations of the possible objects in the format of bounding boxes. The bounding boxes are processed with the non-maximum suppression technique to filter out repetitions. Denote the object proposals as $\{b_j\}_{j=1}^R$ and image regions cropped from corresponding boxes as $\{p_j\}_{j=1}^R$. We first select the indices of the bounding boxes with objects that can be named by our customized tag set,
\begin{equation}
    \mathcal{P} = \{\ell_k\}_{k=1}^Q = \{j | \langle f_\text{v}(p_j), f_\text{t}(t_i) \rangle >\beta, i=1, \cdots, N, j=1, \cdots, R \}.
\end{equation}
Here, $\beta$ is a threshold certifying whether $t_i$ is aligned with $p_j$. Given a set of customized attribute $\{a_i\}_{i=1}^V$, each selected proposal $\ell_k$ from $\mathcal{P}$ is then assigned to an attribute
\begin{equation}
    a^*_{\ell_k} = \argmax_{a_i, i=1, \cdots, V} \langle f_\text{v}(p_{\ell_k}), f_\text{t}(a_i) \rangle,
\end{equation}
and the corresponding tags
\begin{equation}
    \mathcal{O}_{\ell_k} = \{t_i | \langle f_\text{v}(p_{\ell_k}), f_\text{t}(t_i) \rangle >\beta, i=1, \cdots, N \}.
\end{equation}
In addition to the tags and attributes to the bounding boxes, we also use the caption model $c(\cdot)$ to provide some more descriptive texts $\{c(p_{\ell_k})\}_{k=1}^Q$.

In summary, we collect a tag set $\cT$ and a caption $c(I)$ as global descriptions to the image, and a quadruple $(b_{\ell_k}, a^*_{\ell_k}, \cO_{\ell_k}, c(p_{\ell_k}))$ as local descriptions for each selected bounding box.

\subsection{Candidate Synthesis}

We then format the collected visual information into the structured \textit{visual clues}, which can be directly used as the prompt of the language model. Figure \ref{fig:framwork} shows an example of the visual clues. We observe that the information near the end of the prompt will have a more significant influence on the language model output. As the tags $\cT$ are usually more informative and the local extractions are noisier, we input the visual clues with the order of local descriptions, caption, and tags.

To incorporate each local description, a naive way is to inject the coordinates of the bounding boxes directly into the prompt. However, we find the current language models still lack the capability to handle the inference task with numbers, especially in a zero-shot manner. Therefore, we reformat the bounding boxes $b_{\ell_k}$ into plain language by describing its location and size. Specifically, we adopt rule-based method to divide the locations into $9$ classes $\{$``\textit{upper left}'', ``\textit{upper middle}'', ``\textit{upper right}'', ``\textit{left}'', ``\textit{middle}'', ``\textit{right}'', ``\textit{lower left}'', ``\textit{lower middle}'', ``\textit{lower right}''$\}$, and divide the sizes into $3$ classes $\{$``\textit{large}'', ``\textit{moderate-sized}'', ``\textit{small}''$\}$, and incorporate these descriptions into the prompt. 

The other visual clues are inputted straightforwardly in the format as showed Figure \ref{fig:framwork}. The prompt is then fed into a large-scale language model to synthesize $K$ candidate paragraphs $\{T_i\}_{i=1}^K$ full of descriptive details.

\subsection{Candidate Selection}

Finally, we use the vision-language model again, to select the candidate that aligns best with the image, 
\begin{equation}
    S = \argmax_{T_i, i=1, \cdots, K} \langle f_\text{v}(I), f_\text{t}(T_i) \rangle.
\end{equation}
To further rule out the unrelated concepts in $S$, we filter the output again in sentence level. This is because large language models sometimes have hallucination issues, i.e., it might generate unrelated sentences in the paragraphs. For example, a paragraph beginning with ``\textit{A couple is hugging on the beach.}'' is likely to be followed with ``\textit{It's a beautiful day and they're enjoying the sun and each other's company.}'' even if there is no visual clue suggesting the weather.  Therefore, we split it into sentences $(s_1, s_2, \cdots, s_U)$, and use a threshold $\gamma$ to remove the sentences with lower similarities,
\begin{equation} \label{eq:filter}
    T^* = (s_i | \langle f_\text{v}(I), f_\text{t}(s_i) \rangle>\gamma, i=1, \cdots, U). 
\end{equation}
In this way, we obtain the final output $T^*$.

\section{Automatic Evaluation Metric: SPIPE}
\label{sec:metric}
\begin{figure}
  \centering
  \includegraphics[scale=0.7]{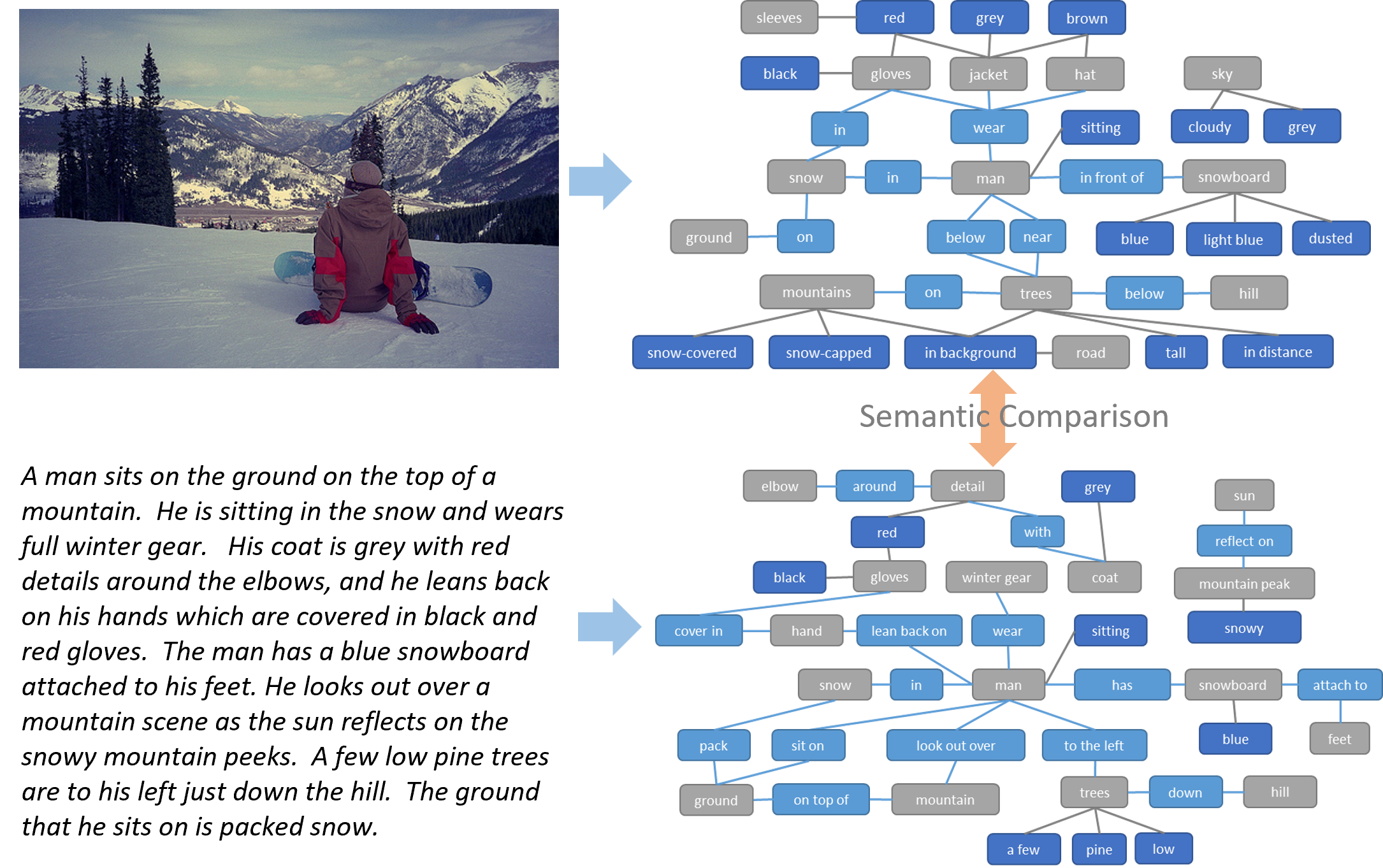}
  \caption{\label{fig:metric} An example of the human-annotated graph and the text extracted graph.}
\end{figure}


As indicated by Figure \ref{fig:illu}, the generated paragraphs of images can be very flexible. This makes the n-gram based metrics, e.g., BLEU \citep{papineni2002bleu}, ROUGE \citep{lin2004rouge}, CIDEr \citep{vedantam2015cider}, METEOR \citep{denkowski2014meteor}, unsuitable for evaluating the generated text. Instead, we focus on the \textit{semantic propositional content}. For example, given an image with content ``\textit{A man sitting in front of a blue snowboard}'', a good evaluation metric for IPC should evaluate whether each of the semantic propositions is correct, namely, a). a man is sitting; b). a man is in front of a snowboard; c). the snowboard is blue, instead of the exact words used in the text. To do so, SPICE \citep{anderson2016spice} extracts the \textit{scene graphs} \citep{johnson2015image} from the generated texts and the reference texts, respectively, and computes an F-score between the graphs. SPICE targets image caption tasks, where there  are usually multiple good references for each image, and the generation is less flexible. However, IPC tasks usually only have one reference \citep{krause2017hierarchical}, which is not enough to evaluate the flexible generation. Therefore, we propose to directly compare the scene graphs extracted from the generated text to human-annotated graphs. Figure \ref{fig:metric} shows an example of the generated graph from text and the human-annotated graph for the image.

Specifically, a scene graph consists of the objects, the attributes of the objects, and the relationships between the objects. To parse the generated text into a scene graph, we use a two-stage approach following \citet{anderson2016spice}. First, we use the pretrained dependency parser \citep{klein2003accurate} to establish the synthetic dependency between the words. Then we map from the dependency trees to scene graphs using a rule-based system \citep{schuster2015generating}. 
Given scene graphs extracted from the text and the human-annotated graphs \citep{krishna2017visual}, our metric computes 
an F-score based on the synonym match\footnote{Tuples are considered to be matched if their lemmatized word forms are equal or if they are found in the same WordNet \citep{miller1995wordnet} synset.} \cite{denkowski2014meteor} between the two graphs among the conjunction of three sets of concepts: (object), (object, attribute), and (object, relationship, subject). Paying homage to \citet{anderson2016spice}, we name our approach \textbf{SPIPE}, Semantic Propositional Image Paragraph Evaluation.

\section{Empirical Analysis}

The basic evaluation of the generated output should include three aspects: 
\begin{enumerate}[wide, labelwidth=!, labelindent=10pt]
\itemsep0em 
    \item Accuracy. Most of the contents appearing in the paragraph should be from the image;
    \item Completeness. Most of the contents appearing in the image should be included in the paragraph;
    \item Coherence. Paragraphs should be more than concatenating the sentences together. 
\end{enumerate}
We evaluate the accuracy and completeness of the generated descriptions using the proposed automatic evaluation metric SPIPE, and do human evaluation to quantify the coherence. We include more examples in Appendix \ref{sec:more_examples} and $500$ randomly sampled outputs in \href{https://luoweizhou.github.io/captioning_output.html}{this file} for readers to perform a qualitative study.

\subsection{Model Specification}

\noindent\textbf{Models.} We adopt Florence-H \citep{yuan2021florence} as the vision-language model, BLIP-large \citep{li2022blip} finetuned on COCO captions dataset \citep{chen2015microsoft} as the captioner with its default setting, and one-stage detector as a general object detector. To be more specific about the detector, we first omit the category information from COCO \citep{chen2015microsoft} datatset and train Dynamic Head \citep{dai2021dynamic} on the bounding boxes only to formulate a class-agnostic object detector. We then use non-maximum suppression (NMS) to select the top 100 object proposals. 

We use GPT-3 \citep{brown2020language} \textit{Davinci-text-001} model as the language model. To enable more difference in the generated candidates, we adopt temperature as $0.8$. We adopt the frequency penalty as $0.5$ and the maximum number of tokens as $100$. 

\noindent\textbf{Customized sets.} To construct a general domain tag set, we collect the most frequently searched 400 thousand queries in Bing Search as the tags $\{t_i\}_{i=1}^N$. We adopt the attribute set of the Visual Genome dataset \citep{krishna2017visual} as the attribute set $\{a_i\}_{i=1}^V$.

\noindent\textbf{Parameters.} We adopt number of tags $M=5$, thresholds $\beta=\gamma=0.2$, and number of candidates $K=40$. Among $K=40$ candidates, half of them are generated without caption information while the remaining half are with them. This is because we notice the caption model sometimes cannot output good captions due to too small bounding boxes. We also remove the bounding boxes that are smaller than $1/400$ of the image sizes. 

\subsection{Automatic Evaluation}

\begin{table}
  \caption{\label{tab:main}Comparison between different methods using SPIPE metric on the test set of the \textit{Stanford dataset} \citep{krause2017hierarchical}.}
  \label{sample-table}
  \centering
  \begin{tabular}{lllll}
    \toprule
    & Name     & F-score     & Precision & Recall \\
        \midrule
    \multirow{4}{*}{Models} & BLIP-large   & 7.6  &   38.0 & 4.4   \\
    & Socratic model & 3.2 & 13.9 & 1.9 \\
    & BEST--general domain &  8.8 & 15.3 & 6.6     \\
    & BEST--VG domain &   \textbf{10.0} & 17.5 & 7.6    \\
    \midrule
    \multirow{1}{*}{Oracle} & BEST with human extracted visual clues & 22.9 & 32.8 & 19.0 \\ 
    \midrule
        \multirow{2}{*}{Annotation} & Stanford dataset &   17.3 & 27.7 & 14.0    \\
    & Concatenation of VG captions &  18.9 & 40.0 & 14.1     \\
    \bottomrule
  \end{tabular}
\end{table}

In this section, we use SPIPE to benchmark the accuracy and completeness of our framework. We evaluate our framework on the test set of \textit{Stanford dataset} \citep{krause2017hierarchical}. The dataset is a subset of Visual Genome (VG) dataset\footnote{We remark that the VG caption data is included in the pretraining data of BLIP model. Therefore we do not claim our framework as a \textit{zero-shot} method, despite that it can handle the images in the wild in a zero-shot way.} \citep{krishna2017visual}, and therefore we can obtain the human-annotated scene graphs from VG as well. We compare the following frameworks.

\noindent\textbf{BLIP} \citep{li2022blip}. This is the BLIP-large model finetuned on COCO captions dataset.

\noindent\textbf{Socratic model} \citep{zeng2022socratic}. We adopt the image captioning code\footnote{\texttt{https://github.com/google-research/google-research/tree/master/socraticmodels}} without alternation.

\noindent\textbf{BEST-general domain}. This is our framework with the customized set listed above.

\noindent\textbf{BEST-VG domain}. With open-vocabulary capability, our framework can adapt to a specific domain. Here, we replace the customized tag set $\{t_i\}_{i=1}^N$ for the local objects as the object set of VG datasets. 

The results are shown in Table \ref{tab:main}. Our general domain framework significantly outperforms the BLIP model and the Socratic model. With the domain specified to VG, the performance is further boosted. 

\begin{figure}[t!]
  \centering
  \fbox{
    \begin{minipage}[t]{0.25\linewidth}
    \includegraphics[width=8\baselineskip]{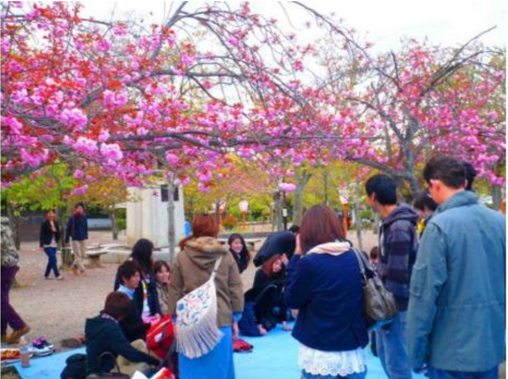}
    \end{minipage}
    \hspace{-10pt}
    \begin{minipage}[t]{0.75\linewidth}
    \vspace{-59pt}
    \textbf{BEST}: \textit{This image features a group of people standing under a tree with pink blossoms. The people are all dressed in various types of clothing, and some are carrying bags or backpacks. The tree is a sakura tree, and the blossoms are in full bloom. There is a bench in the background, and the ground is covered in fallen petals.} 
    
    \textbf{SM}: \textit{People gather under a blossoming cherry tree, enjoying the beauty of natural together.} 
    
    \textbf{BLIP}: \textit{A group of people standing around a blue table.}
    
    \textbf{BEST with only tags}: \textit{The image is of people gathered in a park, looking at cherry blossoms. The parks are full of people enjoying the festival and the beautiful view.}
    \end{minipage}
    }
  \caption{\label{fig:sm} An example cropped from \citet{zeng2022socratic} paper, with other outputs for comparison. }
\end{figure}

Figure \ref{fig:sm} shows an example with a image cropped from the Socratic model paper \citep{zeng2022socratic} directly. We find that caption generation does not require the complex prompt used in Socratic model. Our framework with only tagging information $\mathcal{T}$ can generate texts with a similar degree of detail.

To evaluate the representation capability of our visual clues, we also compare it to a naive scene graph generation method. We use the vision-language model to assign objects, attributes, and relationships between the objects, using the object set, attribute set, and relationship set of VG. And then we compare the generated scene graph to the human-annotated graph. The F-score is $0.3$, with precision $0.8$ and recall $0.2$. We discuss more on why this does not work in Appendix \ref{sec:bad_sgg}.

We also build an oracle model to see the limit of our framework. The oracle model in Table \ref{tab:main} uses the ground truth objects with ground truth attributes to replace the corresponding concepts in the visual clues of our framework. It significantly outperforms the human annotation, either from Stanford dataset or from VG. This reveals the large potential of BEST with the development of object detectors.

\begin{table}[!htb]
\centering
        \caption{\label{tab:ablation}Ablation on each components.  }
  \label{sample-table}
  \begin{tabular}{lllll}
    \toprule
    Name &  F-score     & Precision & Recall \\
    \midrule
    BEST-VG domain &  10.0 & 17.5 & 7.6    \\
    \midrule
    Extraction with YOLOv5 &   9.0 & 19.0 & 6.3   \\
    Remove local information &  8.0 & 14.4 & 6.0    \\
    Remove caption model $c(\cdot)$ &  8.7 & 15.0 & 6.6   \\
    Input tags $\mathcal{T}$ only & 5.9 & 10.7 & 4.4 \\
    Smaller language model (\textit{curie}) & 8.9 & 15.9 & 6.7      \\
    Weaker tagger (CLIP \textit{ViT-L/14}) & 7.8 & 16.4 & 5.6      \\
    \bottomrule
  \end{tabular}
\end{table}

\subsection{Ablation}

We perform an ablation study to see how each of the components contributes to the final performance. Especially, we consider replacing the open-vocabulary object detector with YOLO v5 \citep{yolov5}, which is a closed-set detector trained with COCO classes. Table~\ref{tab:ablation} shows the results. The performance of the YOLO v5 alternation is competitive compared to our general domain version. The precision is higher than ours, which may be a consequence that YOLO models tend to recognize fewer objects \citep{od_20_years}. However, it is still inferior to our VG domain model.

\subsection{Human Evaluation}

To further evaluate our framework, we perform human evaluation. We first compare BEST to human annotation. 
Specifically, we randomly sample $200$ descriptions from the test set of the two sources. For each assignment, we present one image and two corresponding descriptions, and ask human evaluators to evaluate on accuracy, completeness, coherence, and ask an additional question ``\textit{which of the descriptions is written by human}'' for the humanlikeness aspect. They will choose one answer from \{\textit{Description 1, Description 2, Cannot determine}\}. For each assignment, we hire $5$ workers using the Amazon Mechanical Turk platform. 

As the difference between the long texts can be subtle, we perform two statistical tests to see whether the difference is statistically significant. First, we adopt the voted results of the $5$ workers, and perform a binomial test, treating the $200$ outcomes as \textit{i.i.d} samples. The null hypothesis here is 

\textit{Given an image, the probability that human annotation is better than BEST output is $0.5$. }

To further utilize the detailed outcomes for each image, we perform a Mann–Whitney $U$ test with the null hypothesis as

\textit{The number of people that think (the human annotation is better than the BEST output) is similar to the number of people that think (the BEST output is better than the human annotation).}

Table \ref{tab:human} shows the results. There is \textit{no} significant evidence ($p$ value $\approx0.5$) showing human annotation is better than BEST in terms of completeness and humanlikeness. However, BEST still falls behind in terms of accuracy and coherence. The failure cases are usually because the BEST outputs might contain small mistakes that cannot be easily filtered out, mostly from the hallucination of the language model. We show more examples in  Appendix \ref{sec:neg}.

We then compare BEST to BLIP and the Socratic model using similar hypothesis tests. The results show BEST are significantly better than BLIP and Socratic models under most of the metrics ($p$ value $<0.05$).  Note that here accuracy is defined slightly different than the precision used in Table \ref{tab:main}: In human evaluation, providing background information about concepts in the image is not viewed as inaccurate, while in Table \ref{tab:main} it will hurt the precision.

\begin{table}
  \caption{\label{tab:human} Human evaluation. $p$-value 1 is with the binomial test, and $p$-value 2 is with Mann–Whitney $U$ test.  The blue regions in the voted proportion section represent the proportion that the descriptions from the first source are better than the second, while the orange ones represent the second are better than the first. The $1.2\%$ and $0.6\%$ in the middle of row 4 and 12 represent ``\textit{Cannot determine}''.}
  \label{sample-table}
  \centering
  \begin{tabular}{lllcc}
    \toprule
    Sources     & Criteria & Voted proportion     & $p$-value 1  & $p$-value 2\\
    \midrule
    \multirow{4}{*}{Anno. / BEST}  & Accuracy &
    \begin{minipage}{.3\textwidth}
      \includegraphics[width=\linewidth]{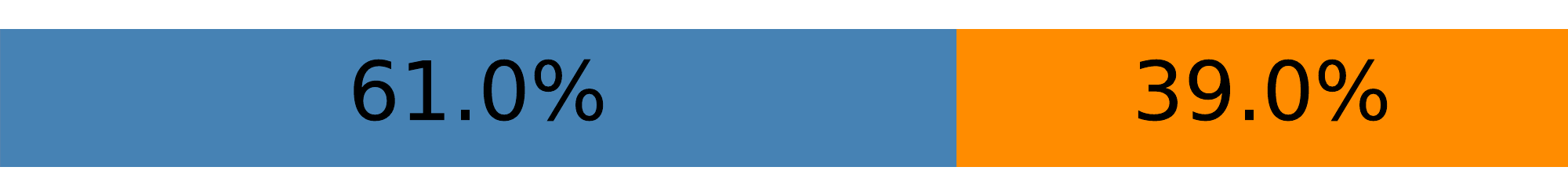}
    \end{minipage}
    & $2\times 10^{-3}$ & $9\times 10^{-6}$
    \\
    & Completeness &
    \begin{minipage}{.3\textwidth}
      \includegraphics[width=\linewidth]{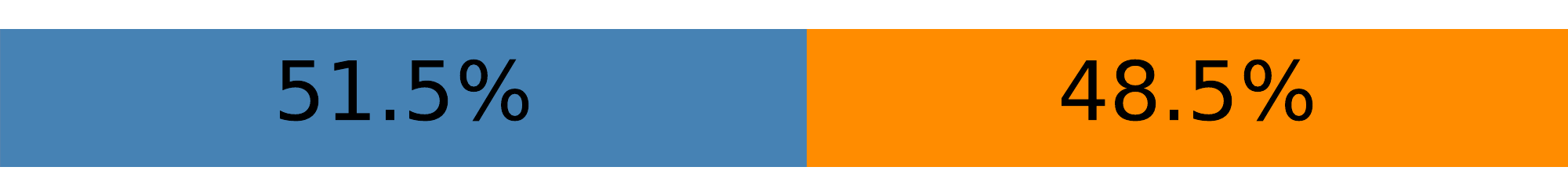}
    \end{minipage}
    & $0.38$ & $0.50$
    \\
    & Coherence &
    \begin{minipage}{.3\textwidth}
      \includegraphics[width=\linewidth]{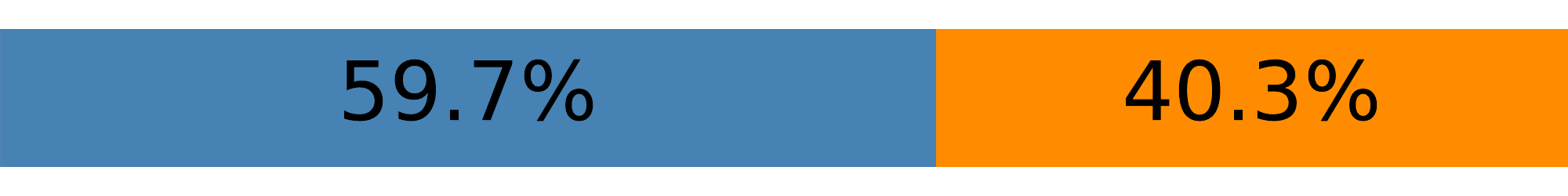}
    \end{minipage}
    & $5\times 10^{-3}$ & $8\times 10^{-4}$
    \\
    & Humanlikeness & \begin{minipage}{.3\textwidth}
      \includegraphics[width=\linewidth]{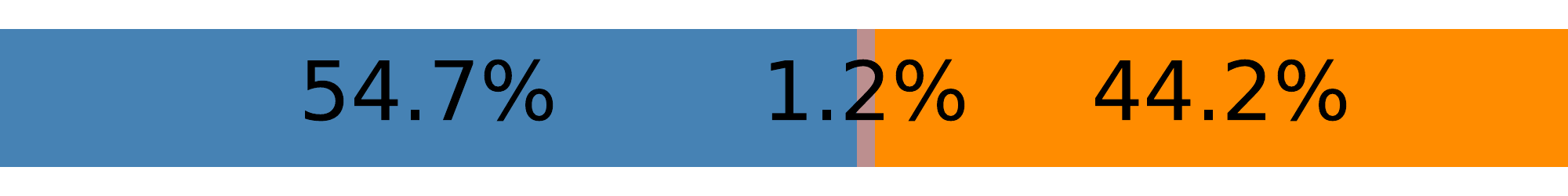}
      \end{minipage}
      & 0.10 & 0.50
    \\
    \midrule
    \multirow{4}{*}{BEST / BLIP}  & Accuracy &
    \begin{minipage}{.3\textwidth}
      \includegraphics[width=\linewidth]{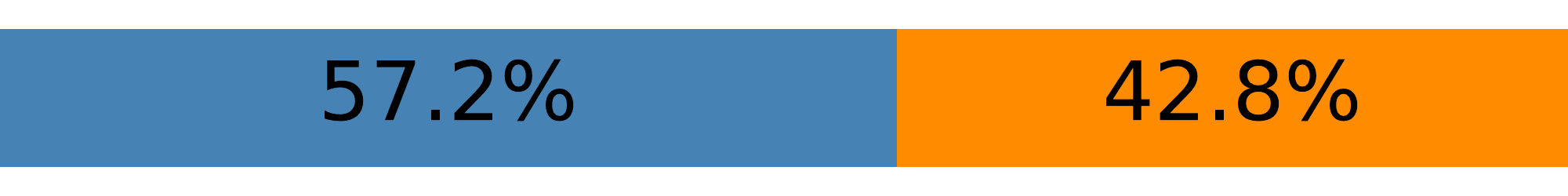}
    \end{minipage}
    & $0.03$ & $3\times 10^{-3}$
    \\
    & Completeness &
    \begin{minipage}{.3\textwidth}
      \includegraphics[width=\linewidth]{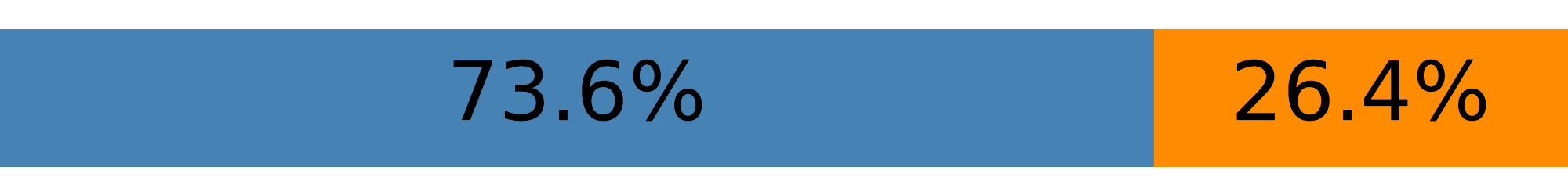}
    \end{minipage}
    & $2\times 10^{-11}$ & $2\times 10^{-28}$
    \\
    & Coherence &
    \begin{minipage}{.3\textwidth}
      \includegraphics[width=\linewidth]{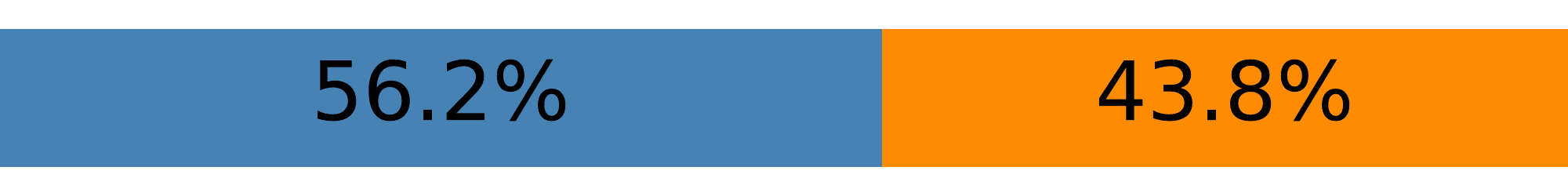}
    \end{minipage}
    & $0.05$ & $5\times 10^{-3}$
    \\
    & Humanlikeness & \begin{minipage}{.3\textwidth}
      \includegraphics[width=\linewidth]{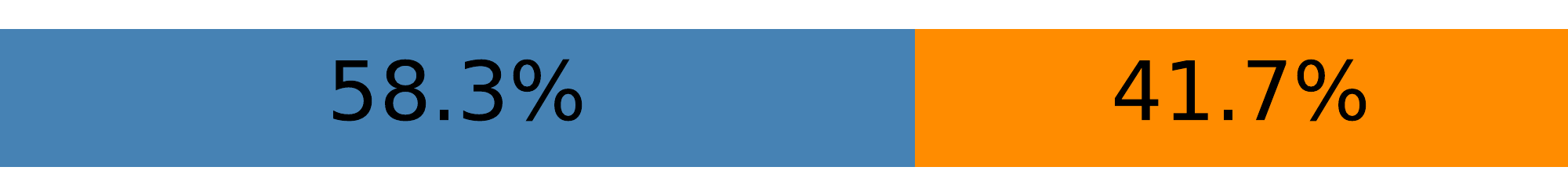}
      \end{minipage}
      & $9\times 10^{-3}$ & $5\times 10^{-4}$
    \\
    \midrule
    \multirow{4}{*}{BEST / Socratic}  & Accuracy &
    \begin{minipage}{.3\textwidth}
      \includegraphics[width=\linewidth]{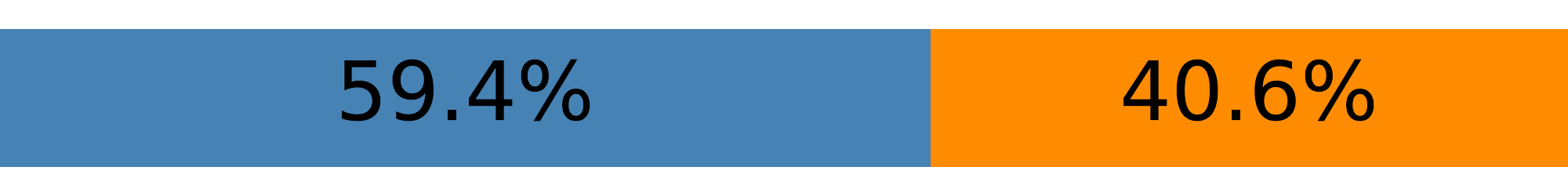}
    \end{minipage}
    & $6\times 10^{-3}$ & $7\times 10^{-5}$
    \\
    & Completeness &
    \begin{minipage}{.3\textwidth}
      \includegraphics[width=\linewidth]{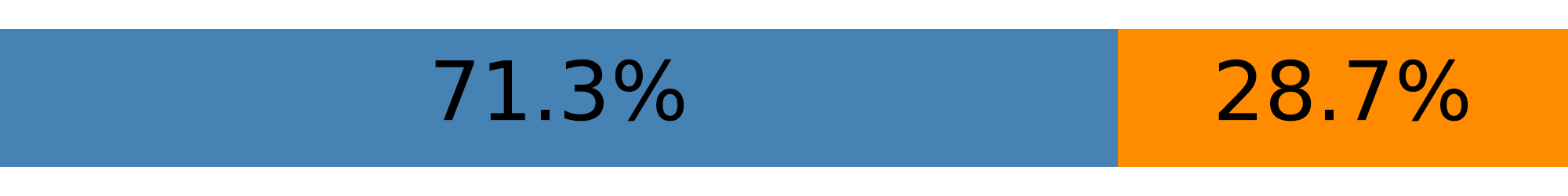}
    \end{minipage}
    & $2\times 10^{-9}$ & $1\times 10^{-24}$
    \\
    & Coherence &
    \begin{minipage}{.3\textwidth}
      \includegraphics[width=\linewidth]{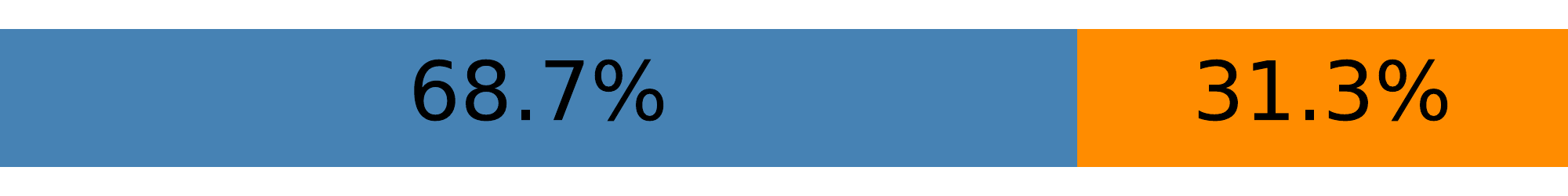}
    \end{minipage}
    & $3\times 10^{-7}$ & $2\times 10^{-12}$
    \\
    & Humanlikeness & \begin{minipage}{.3\textwidth}
      \includegraphics[width=\linewidth]{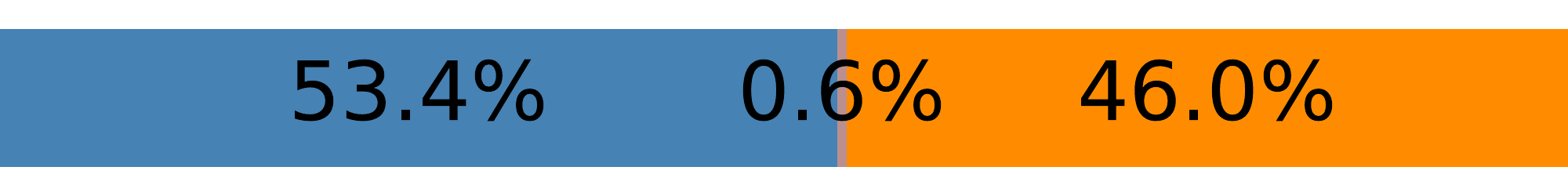}
      \end{minipage}
      & $0.18$ & $0.35$
    \\
    \bottomrule
  \end{tabular}
\end{table}

\section{Variants and Real-world Applications}

The proposed framework opens up many creative real-world applications. For example, people with vision deficiencies may not be able to view images easily. BEST can help convert it into precise and comprehensive descriptions for general domain images.

\begin{table}[!htb]
\centering
        \caption{Finetune BLIP-large on different data.}
  \begin{tabular}{llll}
    \toprule
    Name     & F-score     & Precision & Recall \\
    \midrule
    No finetune & 7.6  &   38.0 & 4.4  \\
    With Socratic model generated data & 3.9 & 17.5 & 2.3 \\
    With BEST generated data & 11.6 & 23.1 & 8.2 \\ 
    \bottomrule
  \end{tabular}

\end{table} 

Another example is the closed-loop training of the large models. 
The large-scale vision-language model and language model used in BEST are trained on tremendous amounts of data, and thus can memorize knowledge beyond human capability. We can use it to automatically annotate data, which is easy to scale up. Furthermore, it can incorporate commonsense knowledge into the text naturally. For example, in the second example of Figure \ref{fig:illu}, the text contains ``\textit{This image captures the everyday life of Cubans, with their traditional horse-drawn carts still in use.}'' This is because our tags contain ``\textit{Cuba}'' and ``\textit{buggy}'', and the language model knows traditional horse-drawn carts are still in use in Cuba.
We finetune a BLIP-large model on our BEST generated data. The training set is similar to Stanford dataset \citep{krause2017hierarchical}. After finetuning, the F-score improves more than $50\%$. 



\begin{figure}
  \centering
  \fbox{
    \begin{minipage}[t]{0.01\linewidth}
    \vspace{-30pt}
    (a)
    \end{minipage}
    \begin{minipage}[t]{0.26\linewidth}
    \centering
    \includegraphics[height=6\baselineskip]{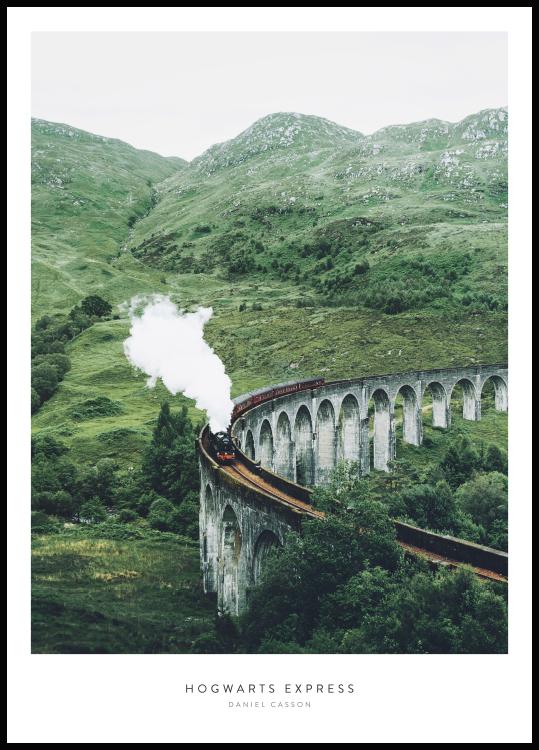}
    \end{minipage}
    \hspace{-10pt}
    \begin{minipage}[t]{0.72\linewidth}
    \vspace{-63pt}
    \textit{As the Hogwarts Express made its way through the Scottish countryside, a young wizard named Harry peered out the window at the stunning Glenfinnan Viaduct. He had always dreamed of traveling on the venerable steam train, and now he was finally doing it! The locomotive chugged along, its great iron wheels clanking over the rails, as passengers in every car gazed in wonder at the beautiful scenery.}
    \end{minipage}
    }
    \fbox{
    \begin{minipage}[t]{0.01\linewidth}
    \vspace{-30pt}
    (b)
    \end{minipage}
    \begin{minipage}[t]{0.26\linewidth}
    \centering
    \includegraphics[height=5\baselineskip]{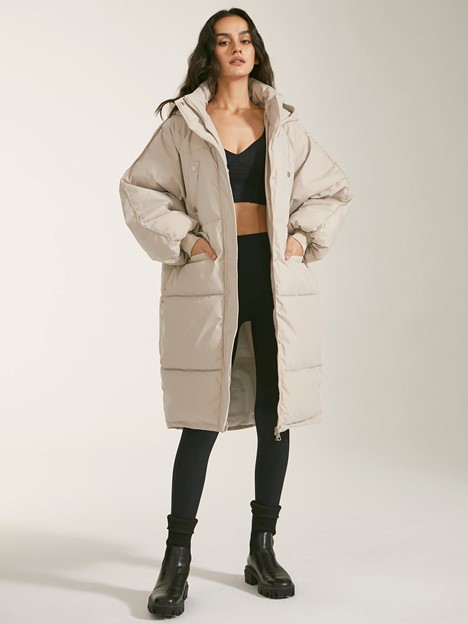}
    \end{minipage}
    \hspace{-10pt}
    \begin{minipage}[t]{0.72\linewidth}
    \vspace{-53pt}
    \textit{Looking for a stylish and practical coat to keep you warm this winter? Look no further than our long parka! Made from high quality beige fabric, this coat is sure to keep you warm and comfortable all winter long. It features a stylish and flattering fit, as well as a warm and cozy hood, making it perfect for cold days out.}
    \end{minipage}
    }
    \fbox{
    \begin{minipage}[t]{0.01\linewidth}
    \vspace{-30pt}
    (c)
    \end{minipage}
    \begin{minipage}[t]{0.26\linewidth}
    \centering
    \includegraphics[height=3.5\baselineskip]{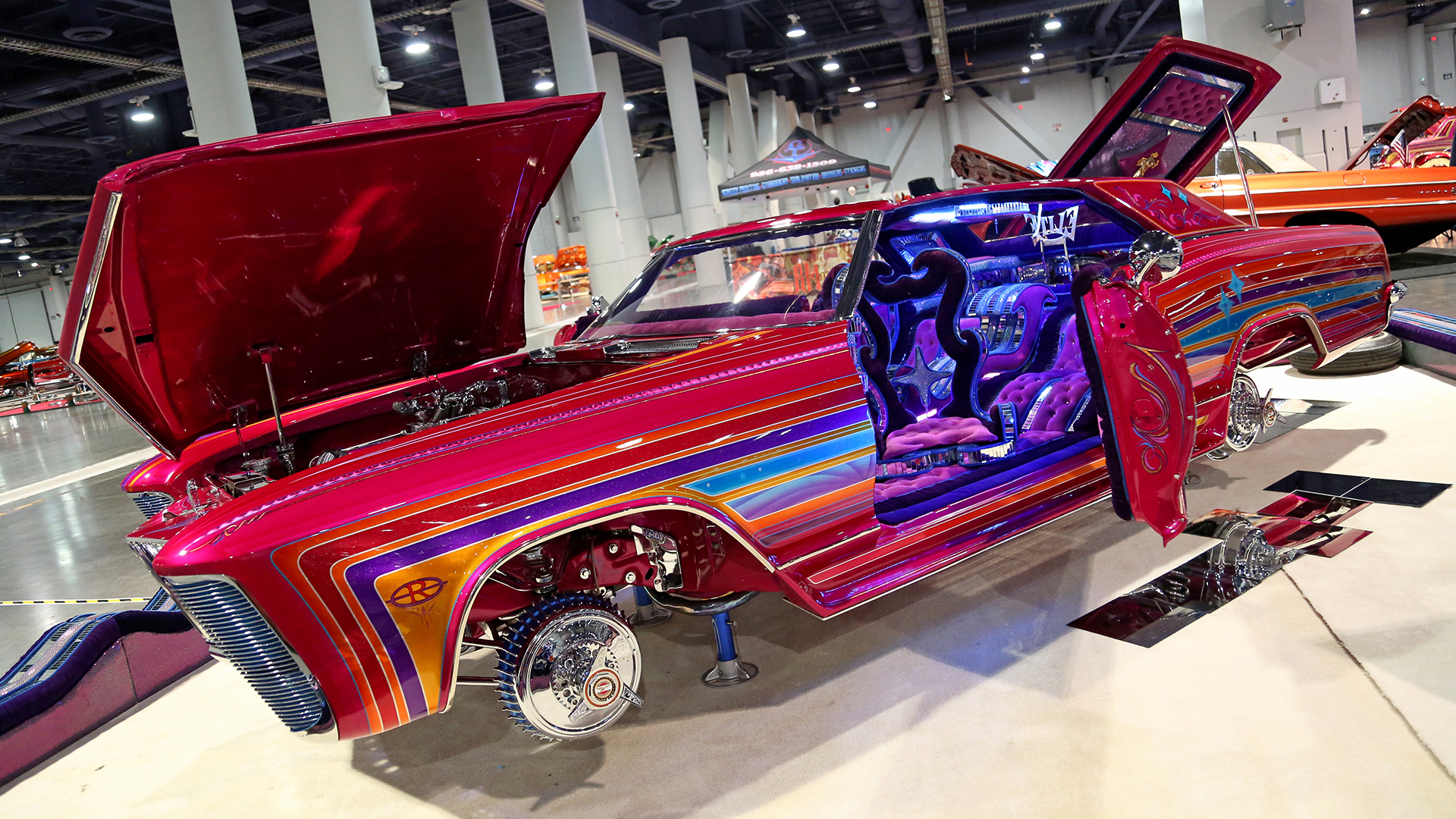}
    \end{minipage}
    \hspace{-10pt}
    \begin{minipage}[t]{0.72\linewidth}
    \vspace{-28pt}
    \textit{What do you think of this Chevy Impala Lowrider? I spotted it at the Sema Car Show and it's definitely a head-turner!}
    \end{minipage}
    }
    \fbox{
    \begin{minipage}[t]{0.01\linewidth}
    \vspace{-25pt}
    (d)
    \end{minipage}
    \begin{minipage}[t]{0.26\linewidth}
    \centering
    \includegraphics[height=4\baselineskip]{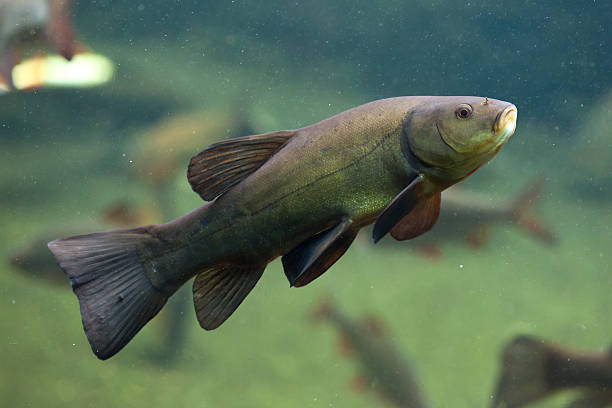}
    \end{minipage}
    \hspace{-10pt}
    \begin{minipage}[t]{0.72\linewidth}
    \vspace{-43pt}
    \textit{A tench is a fish that lives in fresh water. It is a common fish that can be found in many lakes and rivers. The tench is a large fish, and can grow to be quite large. It has a greenish-brown body, and is covered in scales. The tench feeds on insects, plants, and small fish.}
    \end{minipage}
    }
    \fbox{
    \begin{minipage}[t]{0.01\linewidth}
    \vspace{-30pt}
    (e)
    \end{minipage}
    \begin{minipage}[t]{0.26\linewidth}
    \centering
    \includegraphics[height=5\baselineskip]{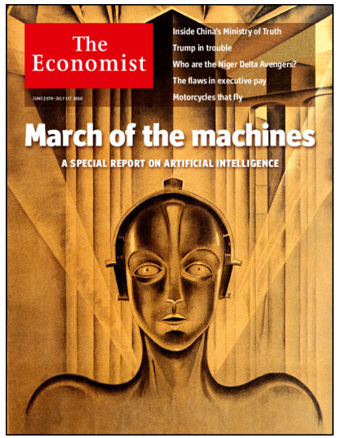}
    \end{minipage}
    \hspace{-10pt}
    \begin{minipage}[t]{0.72\linewidth}
    \vspace{-53pt}
    \textit{The image is of a cover of The Economist magazine. The headline is “March of the machines”. The cover features a drawing of the Maschinenmensch, or machine man, in the center. The background is art deco in style. This image represents the rise of robots and artificial intelligence, as they become increasingly important in our society. }
    \end{minipage}
    }
  \caption{\label{fig:app} Examples for real-world applications with variants of the proposed framework.}
\end{figure}

With small modifications, the proposed framework  enables us to free human labor for even more applications. To list a few examples,

\noindent\textbf{Visual storytelling.} As shown in Figure \ref{fig:app} (a), the framework can generate charming stories based on the input image. To do so, we simply change the end of the prompt to be ``\textit{Tell me a creative story:}''.

\noindent\textbf{Automatic ads generation}: As shown in Figure \ref{fig:app} (b), with the framework, the merchants only need to upload an image, and make small modifications to the generated advertisement as wanted. As there is usually one product in an image, we adopt the number of input tags $M=1$. We also change the end of the prompt to be ``\textit{Write a product description to sell in eBay or Amazon marketplace to get lots of engagement:}''.

\noindent\textbf{Social media post.} As shown in Figure \ref{fig:app} (c), the framework can be a social media bot, which may alleviate the workload of internet celebrities. We change the end to be ``\textit{Social media post:}''.

\noindent\textbf{Background generation.} As shown in Figure \ref{fig:app} (d), the framework can also be used to provide some background knowledge. To do so, we change the end of the prompt to be ``\textit{Textbook text:}''.

\noindent \textbf{Applications with scene texts.} As shown in Figure \ref{fig:app} (e), when the input image has rich scene text, we plug in an OCR (Optical Character Recognition) model \citep{patrickfarley}, and insert the OCR output into the prompts with prefix ``\textit{This image contains text:}'' before the captions\footnote{This variant is featured in June 11th 2022 edition of 
\href{https://www.economist.com/interactive/briefing/2022/06/11/huge-foundation-models-are-turbo-charging-ai-progress}{\textit{The Economist}}. }.

\section{Applications to Visual Question Answering}

The visual clues is a faithful and detailed description of the image, which can be used to answer visual questions leveraging the question answering ability of language models. Specifically, we replaced the ending of the prompt to be the question, \eg, we replace the ``\textit{Describe the image in detail:}'' by ``\textit{What is the man holding?}''. 
We benchmark its performance in two Visual Question Answering (VQA) datasets -- we use the GQA \citep{hudson2019gqa} dataset for probing the capability of scene understanding, and the OK-VQA \citep{marino2019ok} dataset for the awareness of the commonsense knowledge. 

Since no training is performed, BEST generated answer usually have different formatting from the ground truth, causing difficulty in evaluation. For example, for question ``\textit{Is the ground blue or brown?}'', the ground truth answer in GQA is ``\textit{brown}'', but the BEST answer is ``\textit{The ground in the image is brown.}''. Therefore, we use GPT-3 model again to reformat the answer. We refer this evaluation method as \textit{Generative}. Furthermore, for the GQA dataset, the answers in the training set and test set have a large overlap. So we adopt the nearest embedding from the training answers as the final answer, and refer the method as \textit{Discriminative}. More details can be found in Appendix \ref{sec:vqa_eval}.

\begin{table}[!htb]
\centering
        \caption{\label{tab:vqa}The VQA accuracy on GQA and OK-VQA datasets.}
  \begin{tabular}{llcc}
    \toprule
    Method    & Evaluation & GQA & OK-VQA \\
    \midrule
    \multirow{2}{*}{Socratic}  &   Generative & 24.95 & 16.50 \\
    & Discriminative & 26.89 & -- \\
    \multirow{2}{*}{Visual Clues}  &   Generative & 37.00  & \textbf{28.89} \\
    & Discriminative & \textbf{39.93} & -- \\
    \midrule
    \color{gray}{BLIP}  &   \color{gray}{Exact Match} & \color{gray}{47.58} & \color{gray}{43.62 }\\
    \bottomrule
  \end{tabular}

\end{table} 

\begin{figure}
  \centering
  \fbox{
    \begin{minipage}[t]{0.21\linewidth}
    \centering
    \includegraphics[height=5\baselineskip]{figures/concrete.jpg}
    \end{minipage}
    \begin{minipage}[t]{0.23\linewidth}
    \vspace{-53pt}
    \textbf{Question:}\textit{What material is the crosswalk in front of the stores?}
    
    \textbf{GT: }\textit{Concrete}
    
    \textbf{BEST: }\textit{Concrete}
    \end{minipage}
    }
    \fbox{
    \begin{minipage}[t]{0.19\linewidth}
    \centering
    \includegraphics[height=5\baselineskip]{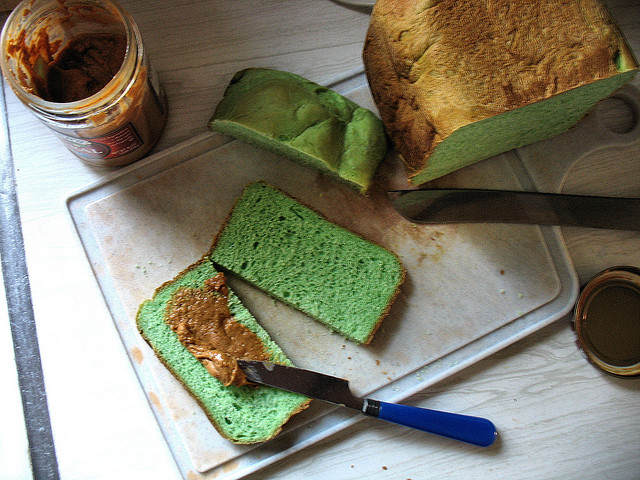}
    \end{minipage}
    \begin{minipage}[t]{0.29\linewidth}
    \vspace{-53pt}
    \textbf{Question:} \textit{What ingredient is missing from the picture to make a pb and j sandwich?}
    
    \textbf{GT: }\textit{Jelly}
    
    \textbf{BEST:} \textit{Jelly}
    \end{minipage}
    }
  \caption{\label{fig:vqa_success} Examples of the success cases from the GQA (left) and the OK-VQA (right).}
\end{figure}

\begin{figure}
  \centering
  \fbox{
    \begin{minipage}[t]{0.21\linewidth}
    \centering
    \includegraphics[height=5\baselineskip]{figures/elephant.jpg}
    \end{minipage}
    \begin{minipage}[t]{0.23\linewidth}
    \vspace{-53pt}
    \textbf{Question:}\textit{What place is pictured?}
    
    \textbf{GT: }\textit{Shore}
    
    \textbf{BEST: }\textit{Africa}
    \end{minipage}
    }
    \fbox{
    \begin{minipage}[t]{0.18\linewidth}
    \centering
    \includegraphics[height=5\baselineskip]{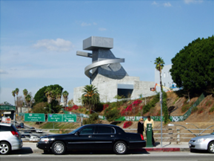}
    \end{minipage}
    \begin{minipage}[t]{0.3\linewidth}
    \vspace{-53pt}
    \textbf{Question:} \textit{Where in the world would you find this structure?}
    
    \textbf{GT: }\textit{California}
    
    \textbf{BEST:} \textit{Los Angeles, California}
    \end{minipage}
    }
  \caption{\label{fig:vqa_failure} Examples of the failure cases from the GQA (left) and the OK-VQA (right).}
\end{figure}

Table \ref{tab:vqa} shows the evaluation results. BEST outperforms Socratic models significantly, suggesting our visual clues are better image representations. We also benchmark the accuracy on BLIP (finetuned on VQA v2 dataset \citep{goyal2017making} and Visual Genome dataset \citep{krishna2017visual}) for reference, which is not directly comparable since its pretrain and finetune datasets have a significant overlap with the evaluation datasets.
Figure \ref{fig:vqa_success} and Figure \ref{fig:vqa_failure} show some success and failure cases of from the VQA datasets.

\section{Limitations and Further Improvements}
\label{sec:limitations}

\noindent\textbf{Prompt tuning.} As suggested in \citet{brown2020language}, language models can infer better when they are shown examples in the prompt. In our experiments, however, this results in model directly copying sentence pieces from example paragraphs, introducing unnecessary noise. We suspect this prompt tuning approach may work better if the input examples are similar to the generated one. This may be a promising direction as we can better control writing style.

\noindent\textbf{Visual clues as a scene graph.} Our visual clue extraction process is motivated by the fact that an image can be comprehensively represented by a scene graph \citep{johnson2015image}. As mentioned in Section \ref{sec:metric}, a scene graph contains objects, attributes of objects, and the relationship between objects. In BEST, however, we do not include the relationships, as we observe in our initial study that  the current vision-language models, although powerful, are not good at inferring relationships (echoing findings from~\citet{thrush2022winoground}). Yet, relationships among the objects are important components of an image. This can be plugged into our framework if better vision-language models are developed.

\noindent\textbf{A well-trained filter model.} We find that the current filtering strategy \eqref{eq:filter} is not immune to certain types of mistakes. As also mentioned in \citet{thrush2022winoground}, the vision-language model cannot accurately associate attributes to their corresponding objects. For example, in the second image of Figure \ref{fig:illu}, if there is a sentence ``\textit{The man wears a green shirt.}'', it 
will lead to a high image-text relevance score, since there is a man, a shirt, and green bush in the image. To handle this issue, we crop the image into local regions and pair each region with an attribute in the clue extraction stage. Still, if it is the language model who hallucinates new attributes and the attributes happen to be in the image, these captions cannot be filtered out. We suspect an adversarially trained filter is needed to perform the task. 

\noindent\textbf{Offensive language, social biases, and stereotypes.} BEST inherits the risks of large vision and language models. It can potentially output offensive language and propagate social biases and stereotypes. For real applications, we can use rule-based methods or train a specific filter to filter out the offensive text. This is an area that we plan to explore to gain more insights further.

\bibliographystyle{ims}
\bibliography{reference}

\newpage

\appendix

\section{More Examples of Generated Paragraphs}
\label{sec:more_examples}
\begin{figure}[h!]
\vspace{20pt}
  \centering
  \fbox{
    \begin{minipage}[t]{0.2\linewidth}
    \includegraphics[width=7\baselineskip]{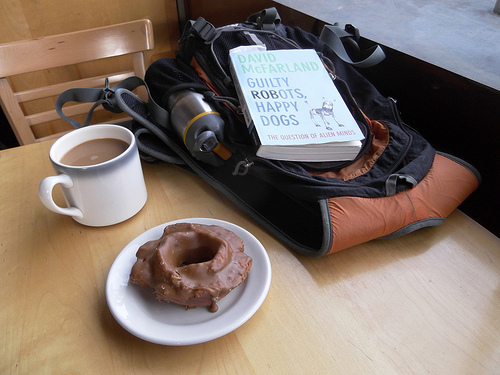}
    \end{minipage}
    \hfill
    \begin{minipage}[t]{0.75\linewidth}
    \vspace{-57pt}
    \textbf{BEST}: \textit{This image features a doughnut and a cup of coffee on a table. The doughnut is chocolate and glazed and the coffee is black. There is also a daypack visible in the image, as well as some straps that may be used to hold the backpack in place.} 
    
    \textbf{SM}: \textit{Home office sink with book and coffee.} 
    
    \textbf{BLIP}: \textit{A doughnut and a cup of coffee on a table.}
    
    \textbf{BEST with only tags}: \textit{This image is of a coffee and donut in a daypack. The donut is labelled the "donut of shame" as it is a bit battered and not as fresh looking as the coffee.}
    
    \textbf{Human Annotation}: \textit{A table has some items on it. The table is made of wood. There is a donut on a plate. The donut is made of chocolate. The plate is white. The plate is round. There is a cup of coffee on the table. The cup is white with a blue rim. The is a backpack by the cup and plate. The backpack is brown and white. There is a book on the backpack. The book is a paperback. The table is by a wall. The table has a chair by it.}
    \end{minipage}
    }
    \fbox{
    \begin{minipage}[t]{0.2\linewidth}
    \includegraphics[width=7\baselineskip]{figures/train.jpg}
    \end{minipage}
    \hfill
    \begin{minipage}[t]{0.75\linewidth}
    \vspace{-42pt}
    \textbf{BEST}: \textit{ This image shows a train traveling over a bridge over a street in Las Vegas. The train is a monorail, and is traveling next to tall buildings and the Stratosphere casino. There is a blinking red light on the train, and a blinking green light on the bridge.} 
    
    \textbf{SM}: \textit{A photo of an indoor parking garage.} 
    
    \textbf{BLIP}: \textit{A train traveling over a bridge over a street.}
    
    \textbf{BEST with only tags}: \textit{This image is of the Las Vegas Monorail. It is a high-speed, driverless railway system that runs on a single beam. The Monorail connects the Las Vegas Strip to the Las Vegas Convention Center and other major hotels, including the Stratosphere, MGM Grand, and Sahara.}
    
    \textbf{Human Annotation}: \textit{There are two skyscrapers. There is also a train going around a track above the road. The traffic lights are lit to red and green. There are palm trees to the left.}
    \end{minipage}
    }
    \fbox{
    \begin{minipage}[t]{0.2\linewidth}
    \includegraphics[width=7\baselineskip]{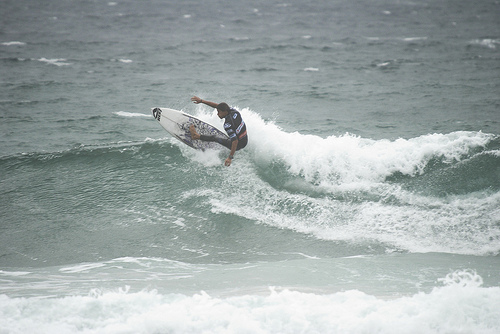}
    \end{minipage}
    \hfill
    \begin{minipage}[t]{0.75\linewidth}
    \vspace{-50pt}
    \textbf{BEST}: \textit{This image captures a surfer performing a cutback maneuver at the Superbank surf competition. The surfer is positioned in the middle of the frame and is relatively small in comparison to the surrounding waves. The waves are large and crashing, providing an impressive backdrop for the surfing action.} 
    
    \textbf{SM}: \textit{A person is surfing on a wave at a medina.} 
    
    \textbf{BLIP}: \textit{A man riding a wave on top of a surfboard.}
    
    \textbf{BEST with only tags}: \textit{This image captures a surfer cutting back in a competition at the Superbank. The surfer is Bruna Surfistinha and the competition is the Chris Ward Surf Invitational. The waves are huge and the surfer is doing a great job of carving through them.}
    
    \textbf{Human Annotation}: \textit{A man is in a black wetsuit. The man is riding a white surfboard. There is a large wave in the water the man is surfing on.}
    \end{minipage}
    }
  \caption{\label{fig:example1} More examples of generated image paragraph.}
\end{figure}

\begin{figure}[h!]
\vspace{20pt}
  \centering
  \fbox{
    \begin{minipage}[t]{0.2\linewidth}
    \includegraphics[width=7\baselineskip]{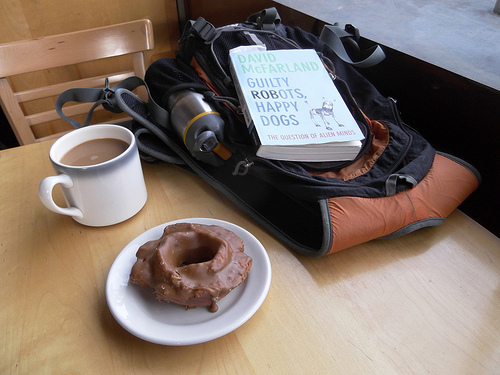}
    \end{minipage}
    \hfill
    \begin{minipage}[t]{0.75\linewidth}
    \vspace{-57pt}
    \textbf{Prompt}: \textit{Objects in this image:}
    
    \textit{a donut on a plate next to a cup of coffee. coffee and donut, is at middle of the image and is large in the image. Attribute: donut}
    
    \textit{a donut on a plate next to a cup of coffee. coffee and donut, is at middle of the image and is large in the image. Attribute: doughnut}

    \textit{a donut on a plate next to a cup of coffee. coffee and donut, is at lower left of the image and is moderate-sized in the image. Attribute: chocolate glazed}

    \textit{Caption: }

    \textit{a doughnut and a cup of coffee on a table}

    \textit{Tags:}
    
    \textit{This image is about coffee and donuts, daypack, the donut of shame, dohnut, randys donuts}

    \textit{Describe this image in detail:}
    \end{minipage}
    }
    \fbox{
    \begin{minipage}[t]{0.2\linewidth}
    \includegraphics[width=7\baselineskip]{figures/train.jpg}
    \end{minipage}
    \hfill
    \begin{minipage}[t]{0.75\linewidth}
    \vspace{-42pt}
    \textbf{Prompt}: \textit{Objects in this image:}

    \textit{a train traveling through a city next to tall buildings. monorail, is at middle of the image and is moderate-sized in the image. Attribute: monorail}

    \textit{a black and white photo of a clock. golden sword, is at lower middle of the image and is small in the image. Attribute: gold-tipped}

    \textit{Caption: }
    
    \textit{a train traveling over a bridge over a street}

    \textit{Tags:}
    
    \textit{This image is about las vegas monorail, las vegas metro, monorail, high roller stratosphere, high roller}

    \textit{Describe this image in detail:}
    \end{minipage}
    }
    \fbox{
    \begin{minipage}[t]{0.2\linewidth}
    \includegraphics[width=7\baselineskip]{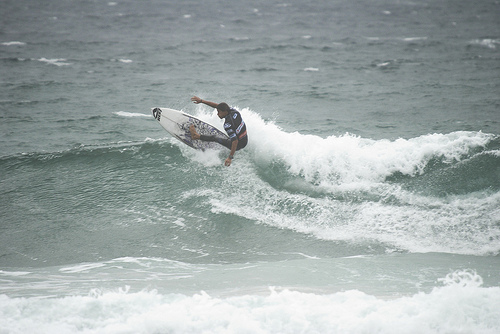}
    \end{minipage}
    \hfill
    \begin{minipage}[t]{0.75\linewidth}
    \vspace{-50pt}
    \textbf{Prompt}: \textit{Objects in this image:}

    \textit{a dog is sitting under an umbrella in the snow. leg on surfboard, is at middle of the image and is small in the image. Attribute: wake boarding}

    \textit{a man riding a wave on top of a surfboard. black surfers, is at middle of the image and is small in the image. Attribute: concentrated surfing}

    \textit{a man riding a wave on top of a surfboard. black surfers, is at middle of the image and is moderate-sized in the image. Attribute: concentrated surfing}

    \textit{a man holding a surfboard on top of a wave. on surfboard, is at middle of the image and is small in the image. Attribute: surfboarding}

    \textit{a man riding a wave on top of a surfboard. man on surfboard, is at middle of the image and is small in the image. Attribute: concentrated surfing}

    \textit{Caption: }

    \textit{a man riding a wave on top of a surfboard}

    \textit{Tags:}

    \textit{This image is about bruna surfistinha, chris ward surfer, superbank, cutback, surf competition}

    \textit{Describe this image in detail:}
    \end{minipage}
    }
  \caption{\label{fig:example1_prompt} Examples of generated prompts of images in Figure \ref{fig:example1}. We can see the prompts may contain minor errors. But language model tends to ignore the concepts that are too irrelevant to the scene. }
\end{figure}

\clearpage
\newpage
\begin{figure}[h!]
  \centering
  \fbox{
    \begin{minipage}[t]{0.2\linewidth}
    \includegraphics[width=7\baselineskip]{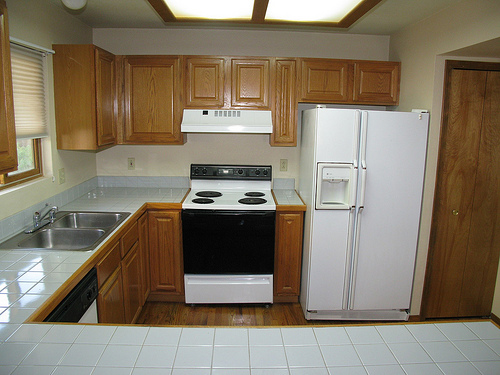}
    \end{minipage}
    \hfill
    \begin{minipage}[t]{0.75\linewidth}
    \vspace{-57pt}
    \textbf{BEST}: \textit{This image shows a kitchen with white appliances and wooden cabinets. The stove is in the middle of the image and the refrigerator is on the right. There is also a sink on the left.} 
    
    \textbf{SM}: \textit{A photo of a kitchen, with a stove, kitchen range, and range hood.} 
    
    \textbf{BLIP}: \textit{A kitchen with white appliances and wooden cabinets.}
    
    \textbf{BEST with only tags}: \textit{ This image is of a kitchen that has been remodeled. It is a galley style kitchen with an oak cabinet. There is a sink, stove, and refrigerator in the kitchen. The countertops are granite and the flooring is wood.}
    
    \textbf{Human Annotation}: \textit{A kitchen that has a brown wooden floor,wooden brown cabinets with a white stove and white refrigerator. the sink is empty and clean along with the counter which is white in color with a window above the sink which has brown blinds.}
    \end{minipage}
    }
    \fbox{
    \begin{minipage}[t]{0.2\linewidth}
    \includegraphics[width=7\baselineskip]{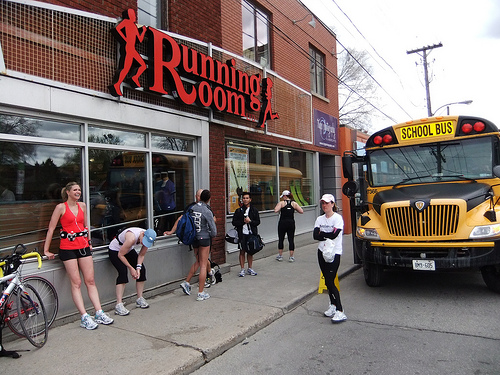}
    \end{minipage}
    \hfill
    \begin{minipage}[t]{0.75\linewidth}
    \vspace{-55pt}
    \textbf{BEST}: \textit{This image is of a group of people who appear to be waiting to enter a running room. The people are standing on a sidewalk and most are wearing athletic clothing. In the background, there is a yellow school bus parked on the side of the road.} 
    
    \textbf{SM}: \textit{Runners in a fastfood restaurant.} 
    
    \textbf{BLIP}: \textit{A group of people standing outside of a running room.}
    
    \textbf{BEST with only tags}: \textit{This image is of a marathon coach and two runners. The marathon coach is giving instructions to the runners and they are getting ready to run. There is a banner in the background that says "Run Around".}
    
    \textbf{Human Annotation}: \textit{A store named the Running Room is to the left of the road. A yellow school bus is parked next to the store. There are people in running clothes lined up the side of the building. Large windows surround the first level of the store. Some bikes are leaned up against the front of the building. There are clouds in the sky and electrical wires are on poles above the road.}
    \end{minipage}
    }
    \fbox{
    \begin{minipage}[t]{0.2\linewidth}
    \includegraphics[width=7\baselineskip]{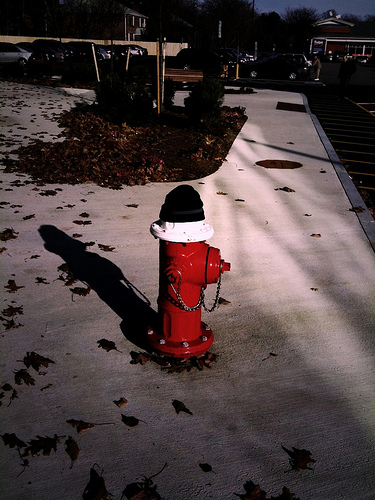}
    \end{minipage}
    \hfill
    \begin{minipage}[t]{0.75\linewidth}
    \vspace{-100pt}
    \textbf{BEST}: \textit{This image is a red fire hydrant that is sitting on the side of the road. The fire hydrant has a black hat on top of it and it is made out of metal. The hydrant is red and the hat is black. The hat might be used to keep the hydrant from freezing in the winter.} 
    
    \textbf{SM}: \textit{The fire hydrant stands out against the snowy background.} 
    
    \textbf{BLIP}: \textit{A red fire hydrant sitting on the side of a road.}
    
    \textbf{BEST with only tags}: \textit{This image shows a firefighter in front of a fire hydrant. The firefighter is wearing a fire hat and has a woodford hydrant. The freeze proof hydrant is behind the woodford hydrant.}
    
    \textbf{Human Annotation}: \textit{A red fire hydrant with a white top and a black knit cap on top of it sits on a sidewalk at the side of a parking lot. The hydrant is casting a shadow on the sidewalk where leaves are scattered around it. There is a manhole cover a bit down from the hydrant. Behind the manhole cover is a small area with small green shrubs and mostly covered by leaves. At the end of the sidewalk are rows of cars with a store in the far back. A couple of people are near their cars in the parking lot. At the end of one row of cars is a wooden fence with a brick building behind it.}
    \end{minipage}
    }
  \caption{\label{fig:example1} More examples of generated image paragraph.}
\end{figure}

\newpage
\begin{figure}[h!]
  \centering
  \fbox{
    \begin{minipage}[t]{0.2\linewidth}
    \includegraphics[width=7\baselineskip]{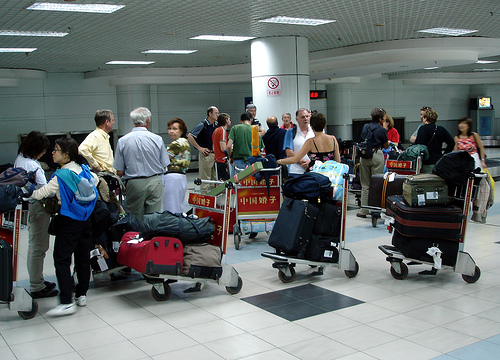}
    \end{minipage}
    \hfill
    \begin{minipage}[t]{0.75\linewidth}
    \vspace{-55pt}
    \textbf{BEST}: \textit{This image shows a group of people gathered around luggage carts at an airport. There is a mix of different types of luggage, including suitcases, backpacks, and duffel bags. There are also a few people in the image who are not carrying any luggage.} 
    
    \textbf{SM}: \textit{People waiting in an airport terminal.} 
    
    \textbf{BLIP}: \textit{A group of people standing around carts with luggage.}
    
    \textbf{BEST with only tags}: \textit{The image shows several people waiting in a line at an airport baggage claim. The airport is likely in China, as both Pudong and Shenzhen airports are located there. Beijing airport is also located in China. All three airports are major international hubs, with flights arriving from all over the world.}
    
    \textbf{Human Annotation}: \textit{There are people standing in a room. There are signs on the walls behind them. There are three scooters that have luggage on top of them in front of the people.}
    \end{minipage}
    }
    \fbox{
    \begin{minipage}[t]{0.2\linewidth}
    \includegraphics[width=7\baselineskip]{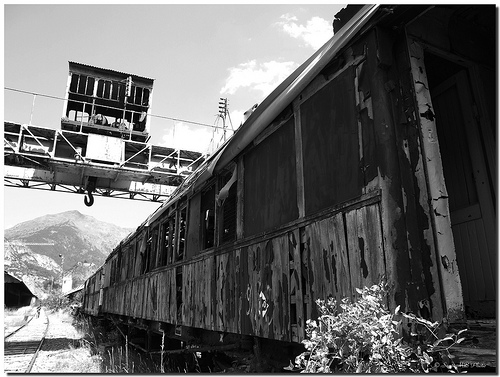}
    \end{minipage}
    \hfill
    \begin{minipage}[t]{0.75\linewidth}
    \vspace{-57pt}
    \textbf{BEST}: \textit{This black and white photo shows an old train station that is now abandoned. The train itself looks quite old and is in the middle of the image, taking up most of the space. There are other objects in the image such as a funicular and cog railroad, but the old train is the main focus.} 
    
    \textbf{SM}: \textit{Abandoned railcar at a forgotten railway platform.} 
    
    \textbf{BLIP}: \textit{A black and white photo of an old train.}
    
    \textbf{BEST with only tags}: \textit{ This image features an abandoned train station. The station is located in the canfranc international railway station, and is part of the durango and silverton narrow gauge railroad. The funiculaire can also be seen in the background, as well as the cog railroad.}
    
    \textbf{Human Annotation}: \textit{A worn wooden building is on the side of a set of train tracks. The building has many windows with screens over the windows. Many of the screens are torn or missing. The paint on the building is severely thinned and peeling. A small foliage bush is at the front of the building. Tall mountains are in the far distance. A large metal structure is suspended in the air above the building and the train tracks.}
    \end{minipage}
    }
    \fbox{
    \begin{minipage}[t]{0.2\linewidth}
    \includegraphics[width=7\baselineskip]{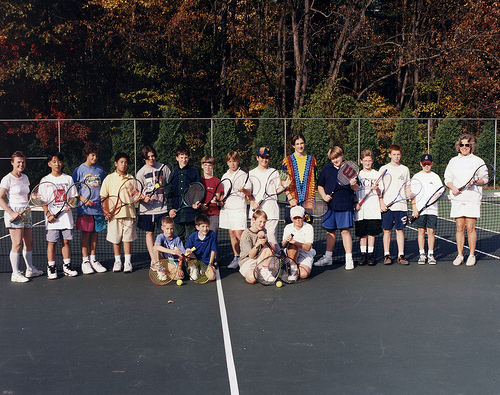}
    \end{minipage}
    \hfill
    \begin{minipage}[t]{0.75\linewidth}
    \vspace{-60pt}
    \textbf{BEST}: \textit{This image features a group of people, all of whom are standing on a tennis court. The people in the image are all dressed in appropriate tennis gear, and each is holding a racquet of some kind. In the background, there is a fence and beyond that, some trees.} 
    
    \textbf{SM}: \textit{A group of people enjoying a game of tennis.} 
    
    \textbf{BLIP}: \textit{A group of people standing on top of a tennis court.}
    
    \textbf{BEST with only tags}: \textit{This image features the Bollettieri Tennis Academy, which is a well-known junior tennis training center. The academy is situated in Bradenton, Florida and has produced many top tennis players, such as Andre Agassi, Monica Seles, and Jim Courier. The academy has a kids' court that is visible in this image, as well as a grand slam tennis court in the background.}
    
    \textbf{Human Annotation}: \textit{ A group of kids and teenagers are posing for a picture on a tennis court. An adult instructor is standing at the far right of the kids. A net is behind them and the court ground is dark green with white painted lines. There are tennis balls on the ground in front of the kneeling kids. A chain link fence is behind the court and a group of trees is in the background. Everyone is wearing shorts, t-shirts and sneakers.}
    \end{minipage}
    }
  \caption{\label{fig:example1} More examples of generated image paragraph.}
\end{figure}

\newpage
\begin{figure}[h!]
  \centering
  \fbox{
    \begin{minipage}[t]{0.2\linewidth}
    \includegraphics[width=7\baselineskip]{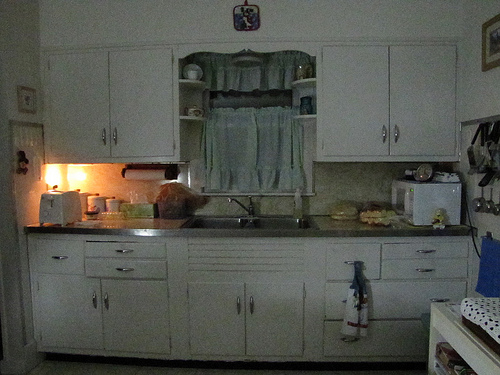}
    \end{minipage}
    \hfill
    \begin{minipage}[t]{0.75\linewidth}
    \vspace{-57pt}
    \textbf{BEST}: \textit{This image is of a kitchen with white cabinets and a sink. The kitchen has a retro feel to it, with the white cabinets and old-fashioned sink. There is a small amount of light in the image, coming from what may be a window in the upper left corner of the photograph. The rest of the image is fairly dark, with only a small amount of light illuminating the countertop and floor.} 
    
    \textbf{SM}: \textit{ A person in a kitchen, cooking food on a stove.} 
    
    \textbf{BLIP}: \textit{A kitchen with white cabinets and a sink.}
    
    \textbf{BEST with only tags}: \textit{This image is of a retro kitchen in the night. It is a small, first kitchen with white cabinets and appliances. There is a light over the sink and a window with a curtain.}
    
    \textbf{Human Annotation}: \textit{White cabinets are in a kitchen. There is a window above the sink with white curtains. There is a white microwave on the counter.}
    \end{minipage}
    }
    \fbox{
    \begin{minipage}[t]{0.2\linewidth}
    \includegraphics[width=7\baselineskip]{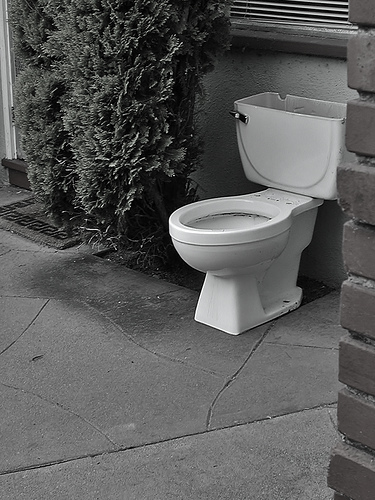}
    \end{minipage}
    \hfill
    \begin{minipage}[t]{0.75\linewidth}
    \vspace{-100pt}
    \textbf{BEST}: \textit{This is a black and white photo of a toilet. The toilet is white and it looks like it is sitting outside of a building next to a bush.} 
    
    \textbf{SM}: \textit{A photo of a toilet.} 
    
    \textbf{BLIP}: \textit{A toilet sitting outside of a building next to a bush.}
    
    \textbf{BEST with only tags}: \textit{This image is of a toilet throne. It is a Universal Rundle toilet with a Portland Loo in the background. The American Standard toilet is to the right of the Portland Loo. The Toto Drake is to the right of the American Standard toilet.}
    
    \textbf{Human Annotation}: \textit{A white toilet with no lid is sitting outside of a brick building on the concrete ground. The toilet is white and is placed against the brick wall of a building. It is in a corner. A tree is on one side of the toilet. A window is in the wall just above the toilet. There are cracks in the concrete on the ground. A door with a doormat is beside the tree.}
    \end{minipage}
    }
    \fbox{
    \begin{minipage}[t]{0.2\linewidth}
    \includegraphics[width=7\baselineskip]{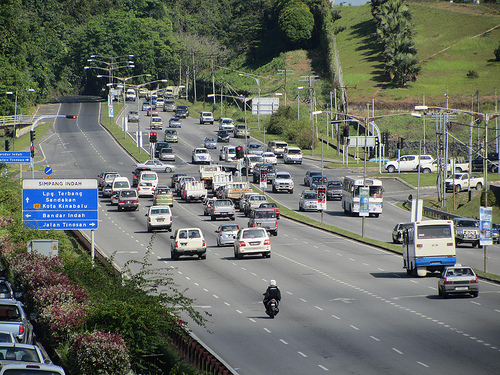}
    \end{minipage}
    \hfill
    \begin{minipage}[t]{0.75\linewidth}
    \vspace{-57pt}
    \textbf{BEST}: \textit{This image captures the traffic on the Subic-Clark-Tarlac Expressway in the Klang Valley region of Malaysia. The expressway is a 4-lane, controlled-access highway that connects the two major metropolitan areas of the country. The image shows several cars and trucks travelling on the highway, as well as an ambulance and a police motorcycle.} 
    
    \textbf{SM}: \textit{A divided highway with many cars and people.} 
    
    \textbf{BLIP}: \textit{A busy city street filled with lots of traffic.}
    
    \textbf{BEST with only tags}: \textit{This image is of the Tarlac-Pangasinan-La Union Expressway (TPLEX), a toll expressway in the Philippines. The expressway is seen here in the Klang Valley area of Selangor, Malaysia, and is crowded with cars and carreolas (two-wheeled carts).}
    
    \textbf{Human Annotation}: \textit{This is a photo of a busy roadway. Many different vehicles can be seen on the road. A long narrow grass strip is separating the different sides of the roads. Tall lamp poles are lining the roadway. A large blue and white sign is standing on the left side of the roadway. Tall green trees and a long steep hill can be seen in the background of the photo.}
    \end{minipage}
    }
  \caption{\label{fig:example1} More examples of generated image paragraph.}
\end{figure}

\newpage

\section{Other Automatic Evaluation Metrics}

In Table \ref{tab:other_metric} we compare different zero-shot methods using the metrics inherit from image captioning tasks. We can see that our framework BEST still performs the best.

\begin{table}[h]
  \caption{\label{tab:other_metric}Comparison between different methods using captioning metrics on the test set of the \textit{Stanford dataset} \citep{krause2017hierarchical}.}
  \label{sample-table}
  \centering
  \resizebox{\columnwidth}{!}{
  \begin{tabular}{llllllll}
    \toprule
    Name     & Bleu-1 & Bleu-2 & Bleu-3 & Bleu-4 & METEOR & ROUGE-L & CIDEr \\
        \midrule
     BLIP-large   & 0.4 & 0.2 & 0.1 & 0.1 & 5.4 & 15.1 & 0.6     \\
     Socratic model &  0.6  & 0.2 & 0.1 & 0.0 & 3.0 & 9.3 & 0.5 \\
     BEST--general domain &  30.9 & 15.9 & 8.1 & 4.2 & 11.9 & 23.7 & 11.3     \\
     BEST--VG domain &  31.8 & 16.5 & 8.5 & 4.4 & 12.5 & 24.3 & 11.9  \\
    \bottomrule
  \end{tabular}
  }
\end{table}

\section{Negative Examples}

\label{sec:neg}

\begin{figure}[h]
  \centering
  \fbox{
    \begin{minipage}[t]{0.3\linewidth}
    \centering
    \includegraphics[width=7\baselineskip]{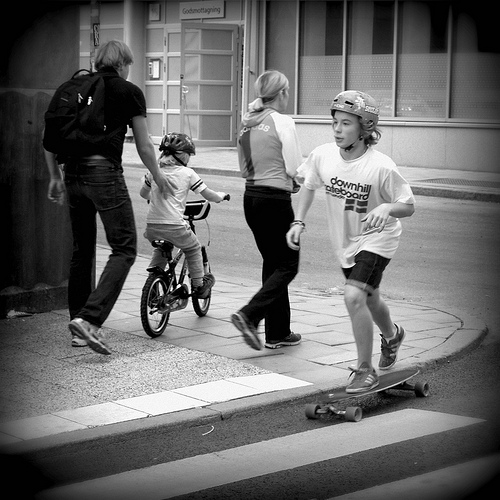}
    \end{minipage}
    \hspace{-10pt}
    \begin{minipage}[t]{0.65\linewidth}
    \vspace{-76pt}
    \textit{In this image, we see a young boy \textbf{riding a bicycle with a backpack on}. To the right, we see another boy skateboarding. In the background, we see people walking and running. \textbf{This image is about streetsurfing, street surfing, street sesh, skatejoring, and longboarding.}}
    \end{minipage}
    }
    \fbox{
    \begin{minipage}[t]{0.3\linewidth}
    \centering
    \includegraphics[width=7\baselineskip]{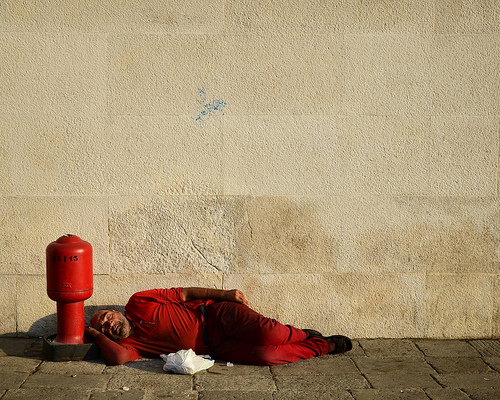}
    \end{minipage}
    \hspace{-10pt}
    \begin{minipage}[t]{0.65\linewidth}
    \vspace{-60pt}
    \textit{This image captures a homeless man taking a much-needed nap in the middle of the day. The bright red fire hydrant is a stark contrast to his muted red clothing, and stands out in the image. The man is lying on his back, with his arms and legs sprawled out, and appears to be asleep. He is surrounded by empty bottles and cans, as well as a few pieces of trash. \textbf{There is an inscription on the wall behind him that reads "Siesta".}}
    \end{minipage}
    }
    \fbox{
    \begin{minipage}[t]{0.3\linewidth}
    \centering
    \includegraphics[width=7\baselineskip]{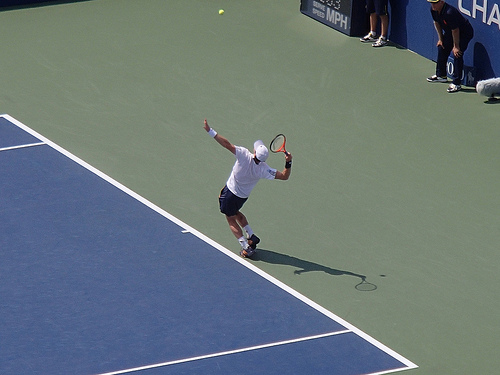}
    \end{minipage}
    \hspace{-10pt}
    \begin{minipage}[t]{0.65\linewidth}
    \vspace{-57pt}
    \textit{This image features Andy Roddick serving during the US Open Tennis Championships in 2009. Roddick is a former world number one tennis player and is known for his powerful serve. In the image, he is shown hitting a forehand shot. \textbf{The ball is seen speeding towards the opponent's court, and the racquet appears to be at the top of its arc.} The umpire can be seen in the background, and the top of the racquet is visible in the foreground.}
    \end{minipage}
    }
  \caption{\label{fig:neg} Negative examples of generated image paragraph.}
\end{figure}

Figure \ref{fig:neg} shows a few failure cases of BEST generation, which may highlight the direction for further improvement. The first example demonstrates two issues. The first one is the vision-language model cannot accurately
match attributes to the objects, as we discussed in Section \ref{sec:limitations}. The second issue is the language model sometimes repeat the input sentences. This issue will be alleviated if we adopt better language models, e.g., replacing \textit{Davinci-text-001} by its upgraded version \textit{Davinci-text-001}. 

The second example suggests the Optical Character Recognition (OCR) capability of the vision-language model is not well-trained. So if for datasets with images involving text photos, we should explicitly plug in an OCR model. 

The third example shows the current VL model cannot differentiate the relationship between the objects. This may be because the training loss, e.g., contrastive loss, does not require the model to differentiate the complex relationships -- only recognizing the objects is enough.

\section{Scene Graph Generation Baseline}
\label{sec:bad_sgg}

As mentioned earlier, to evaluate the representation capability of our visual clues, we also compare it to a naive scene graph generation method. We use the vision-language model to assign objects, attributes, and relationships between the objects, using the object set, attribute set, and relationship set of VG. Algorithm \ref{alg:sgg} shows the complete scene graph generation algorithm. And then we compare the generated scene graph to the human-annotated graph. 

\begin{algorithm}
\caption{Algorithm for the Scene Graph Generation Baseline}\label{alg:sgg}
\begin{algorithmic}
\STATE Select the bounding boxes in the same way as BEST
\STATE Associate each of the bounding boxes with one object name and one attribute
\FOR{each pair of bounding boxes}
\STATE Compute the minimum bounding box covering the union of the two bounding boxes
\STATE Assign a relationship to the minimum bounding box using the open-vocabulary tagger
\STATE Choose from (object 1, relationship, object 2) and (object 2, relationship, object 1) with tagger
\STATE Add the chosen relationship with its object and subject to the output
\ENDFOR
\end{algorithmic}
\end{algorithm}

\begin{figure}[htb!]
  \centering
  \fbox{
    \begin{minipage}[t]{0.2\linewidth}
    \includegraphics[height=4.5\baselineskip]{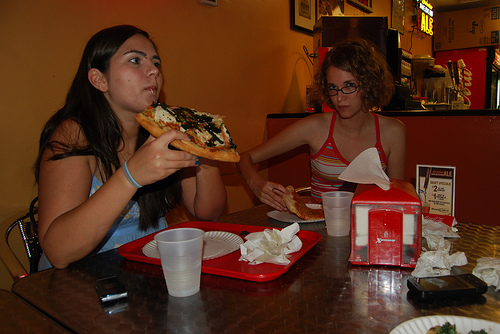}
    \end{minipage}
    \begin{minipage}[t]{0.75\linewidth}
    \vspace{-45pt}
    \textbf{Model Generation} \\
    \textbf{Objects}: \textit{plastic cups, woman eats pizza, spinach on pizza, enjoying a sandwich, eating contest, plastic cup, ribs, case is brown, griddle, foot pedal, parsley on plate }\\
    \textbf{Attributes}: \\
    (object: \textit{plastic cups}, attr: \textit{solo cup}), \\
    (object: \textit{woman eats pizza}, attr: \textit{eating italian}), \\
    (object: \textit{spinach on pizza}, attr: \textit{looking at pizza}), \\
    (object: \textit{enjoying a sandwich}, attr: \textit{getting her lunch}), \\
    (object: \textit{plastic cup}, attr: \textit{solo cup}), \\
    (object: \textit{ribs}, attr: \textit{bacon}) ... \\
    \textbf{Relationships}: \\
    (object: \textit{woman eats pizza}, subject: \textit{plastic cups}, rel: \textit{eating pizza}), \\
    (object: \textit{spinach on pizza}, subject: \textit{plastic cups}, rel: \textit{eating pizza}),\\
    (object: \textit{plastic cups}, subject: \textit{enjoying a sandwich}, rel: \textit{eating pizza}),\\
    (object: \textit{case is brown}, subject: \textit{plastic cups}, rel: \textit{cup on}), \\
    (object: \textit{griddle}, subject: \textit{plastic cups}, rel: \textit{table has items})... \\
    \rule{\columnwidth}{0.01cm}\hfill
    \textbf{Human Annotation} \\
    \textbf{Objects}: \textit{"bracelet", "glasses", "face", "cup", "phone", "table", "pizza", "hand", "tray", "neon sign", "wall", "napkin holder", "picture", "paper plate", "women", "pizza", "cups", "paper napkins", "woman", "tanktop", "woman", "blue shirt", "napkin", "plate", "some food", "advertisement", "cell phone", "short hair", "white napkins", "toppings", "ale", "plate and napkin", "dark hair", "wrist" }\\
    \textbf{Attributes}: \\
    (object: \textit{bracelet}, attr: \textit{blue}), \\
    (object: \textit{glasses}, attr: \textit{black}), \\
    (object: \textit{cup}, attr: \textit{plastic}), \\
    (object: \textit{pizza}, attr: \textit{large}), \\
    (object: \textit{cell phone}, attr: \textit{black}) ... \\
    \textbf{Relationships}: \\
    (object: \textit{face}, subject: \textit{glasses}, rel: \textit{on}), \\
    (object: \textit{table}, subject: \textit{phone}, rel: \textit{on}),\\
    (object: \textit{hand}, subject: \textit{pizza}, rel: \textit{in}),\\
    (object: \textit{table}, subject: \textit{tray}, rel: \textit{on}), \\
    (object: \textit{wall}, subject: \textit{picture}, rel: \textit{on})...
    \end{minipage}
    }
  \caption{\label{fig:neg_sgg} One examples of the generated scene graphs.}
\end{figure}

The F-score is 0.3, with precision 0.8 and recall 0.2. We show an example of the output graphs in Figure \ref{fig:neg_sgg}. The particularly bad performance is majorly due to the sets from VG are noisy, so the synonym match between nodes yields very low score.

\section{More Details on VQA Evaluation}
\label{sec:vqa_eval}

For generative evaluation, we show two made-up examples to GPT-3, to let it know the level of detailness we are looking for. For example, for the 
question ``\textit{Is the ground blue or brown?}'' and the BEST answer  ``\textit{The ground in the image is brown.}'', we feed the following text into GPT-3:

\textit{Question: What is this bird called?}

\textit{Long answer: The bird in this image is called a cockatoo.}

\textit{Short answer: Cockatoo.}

\vspace{10pt}

\textit{Question: Is the chair on the left or on the right of the desk?}

\textit{Long answer: The chair is on the left of the desk. }

\textit{Short answer: Left.}

\vspace{10pt}

\textit{Question: Is the ground blue or brown? }

\textit{Long answer: The ground in the image is brown.}

\textit{Short answer:}

The two examples are fixed for all the evaluations.

For discriminative evaluation, we collect the possible answers $\{a^{\text{t}}_{i}\}$ in the training set, and encode them with text encoder $f_\text{t}(\cdot)$. We then encode the generative short answer $a^{\text{g}}$ generated as above, and adopt the answer with highest similarity as the final answer $a^{\text{f}}$,
\begin{align*}
    a^{\text{f}} = \argmax_{a^{\text{t}}_{i}} \langle f_\text{t}(a^{\text{t}}_{i}), f_\text{t}(a^{\text{g}}).
\end{align*}

\end{document}